\documentclass[sigconf]{acmart}

\AtBeginDocument{%
  \providecommand\BibTeX{{%
    \normalfont B\kern-0.5em{\scshape i\kern-0.25em b}\kern-0.8em\TeX}}}

\setcopyright{acmcopyright}

\acmConference{ArXiv}{Preprint}{2022}

\usepackage{multicol}
\usepackage{multirow}
\usepackage{subcaption}

\usepackage{hyperref}
\usepackage{url}
\usepackage{microtype}
\usepackage{booktabs} 

\usepackage{graphicx}
\usepackage{epsfig}
\usepackage{amsmath}
\usepackage{amsfonts}
\usepackage{graphicx}
\usepackage{comment}
\usepackage{wrapfig}
\usepackage{placeins}

\newcommand{\be}{\begin{equation}}
\newcommand{\ee}{\end{equation}}

\newcommand{\bea}{\begin{eqnarray}}
\newcommand{\eea}{\end{eqnarray}}

\newcommand{\bi}{\begin{itemize}}
\newcommand{\ei}{\end{itemize}}

\newcommand{\ben}{\begin{enumerate}}
\newcommand{\een}{\end{enumerate}}

\newcommand{\bef}{\begin{figure*}}
\newcommand{\enf}{\end{figure*}}

\newcommand{\bt}{\begin{tabular}{lcllcl}}
\newcommand{\et}{\end{tabular}}

\newcommand{\bd}{\begin{description}}
\newcommand{\ed}{\end{description}}

\newcounter{example}

\newcommand{\eref}[1]{(\ref{#1})}       

\newcommand{\dfn}{\stackrel{\triangle}{=}}  

\newcommand{\avec} {{\mathbf a}}

\newcommand{\wvec} {{\mathbf w}}
\newcommand{\xvec} {{\mathbf x}}

\newcommand{\bvec} {{\mathbf b}}

\newcommand{\D}{\Delta}

\begin{document}

\title{Real World Large Scale Recommendation Systems Reproducibility and Smooth Activations}

\author{Gil I. Shamir, Dong Lin}
\email{gshamir@google.com, dongl@google.com}
\affiliation{Google \country{}}
\renewcommand{\shortauthors}{Shamir and Lin}


\begin{abstract}
Real world recommendation systems influence a constantly growing set of domains.  With deep networks, that now drive such systems, recommendations have been more relevant to the user's interests and tasks. However, they may not always be reproducible even if produced by the same system for the same user, recommendation sequence, request, or query.  This problem received almost no attention in academic publications, but is, in fact, very realistic and critical in real production systems. We consider reproducibility of real large scale deep models, whose predictions determine such recommendations.  We demonstrate that the celebrated Rectified Linear Unit (ReLU) activation, used in deep models, can be a major contributor to irreproducibility.  We propose the use of smooth activations to improve recommendation reproducibility.  We describe a novel family of smooth activations; Smooth ReLU (\emph{SmeLU}), designed to improve reproducibility with mathematical simplicity, with potentially cheaper implementation.  SmeLU is a member of a wider family of smooth activations.
While other techniques that improve reproducibility in real systems usually come at accuracy costs, smooth activations not only improve reproducibility, but can even give accuracy gains.  We report metrics from real systems in which we were able to productionalize SmeLU with substantial reproducibility gains and better accuracy-reproducibility trade-offs.  These include  click-through-rate (CTR) prediction systems, content, and application recommendation systems.
\end{abstract}

\keywords{Deep networks, reproducibility, nondeterminism, underspecification, click-through-rate prediction, recommendation systems.}

\maketitle

\section{Introduction}
\label{sec:introduction}
Machine learned recommendation systems are appearing in more and more domains.  Examples include systems for search engines, content recommendation, online shopping, online ads, and diagnostic.  Deep models, that have grown to dominate and drive such systems, can recommend how to find answers to questions, what movie to watch, what product to buy, which ads are shown, how to fix one's car, and how to diagnose one's medical condition.  Similar models are the backbone of language translation and understanding, speech recognition, image understanding, video processing, and other applications. 

Consider a situation where a user queried for a job recommendations and two attractive results were shown.  The user chose one, and deferred the other for later, without saving the details.  Later, the user queries again to find the second job recommendation, but the system showes a different recommendation list.  The user is unable to find the job they want.  This scenario is very realistic and common.  An identical query to the same system gives a different set of recommendations, and no matter how hard one tries, they cannot reproduce a result set they had found before.  For the user, this can be frustrating.  
There may be serious consequences in other applications, like medical diagnosis. \emph{Irreproducibility\/} in deep models is a critical factor in causing situations like this. Unlike the crisis of \emph{lack of replicability\/} \cite{pineau2021improving} in published results due to missing information, much less attention has been given to this problem.

While deep models provide astonishing results on many tasks, they appear to unveil model irreproducibility.
The problem manifests as 
multiple models with the same configuration and training data generate predictions with deviations far beyond classical statistical deviations. The scenario we described is one example. This problem has become very obvious in practical large scale systems.  Despite its importance, it received very little attention in academic publications.  Only recently, a series of empirical works \cite{chen20,damour20,dusenberry20,shamir20,shamir20ed,snapp2021,summers21,yu2021dropout} demonstrated it.
An initial theoretical framework for reproducibility in optimization only appears in very recent work \cite{ahn21reproducibility} and demonstrates the problem for the much simpler case of convex optimization. Generalizing such results to deep learning, with highly non-convex loss landscape,  is even more challenging.

In real systems, two models that are supposedly the same, may behave very differently when deployed in production.
For some applications this may be acceptable.  However, for some recommendation systems, providing different recommendations for the same request is undesirable, specifically in cases where a typical user would expect equal results.
Irreproducibilty is exacerbated if the current recommendations and their user engagements become the training data for future recommendations, as in reinforcement learning systems.  Divergence of the training data gradually expands model and future recommendation differences.  Practical systems often retrain models and redeploy new versions, use continuous (online) training, update models with fresh new data, or replace models by new generations.  Different recommendations may be unavoidable if new generations or new data change model predictions, but one would expect fresh versions or a small amount of new data not to cause substantial deviations.

Irreproducibility affects not only the user, but also the engineering development cycle. It can lead to arbitrary model evaluations, and invalid A/B testing results. 
Model developers usually use aggregate prediction accuracy metrics on either validation or progressive validation \cite{blum99} (in online models) data to judge experimental models. With huge training datasets, training is expensive, timely, and resource intensive.  Model developers attempt to fail fast based on training performance, to only deploy promising models for live recommendations. 
Without taking reproducibility into consideration, 
a model can be deployed in experimental stage, show favorable metrics, but then, when retrained to be deployed in production, produce unfavorable results. This can be expensive to diagnose and address. An example of large scale recommendation is click-through-rate (CTR) prediction in sponsored advertising \cite{mcmahan13}.  The signal produced by the model is used as part of an auction process to determine ads to show.  Shown ads choices affect other system metrics.  Two models that are irreproducible tend to show significant divergence in these system metrics, and also end up pushing different examples to the training data. 
Irreproducibility is thus a source of concern to the engineering development cycle.  The problem is not unique to CTR systems, and can interfere with development cycles of other recommendation systems in a similar fashion.

Unlike overfitting, duplicate models that produce different recommendations may have identical average loss or accuracy metrics, but they can exhibit potentially very large \emph{Prediction Differences} (PDs) on individual examples \cite{chen20, dusenberry20, yu2021dropout}. PDs do not diminish with more training examples (unlike epistemic/model uncertainty). We may observe PDs in orders of magnitude of 20\% or more of the predicted positive engagement rates.  Specifically, we observed that PDs are very correlated to system irreproducibility.  Thus they can be used to measure the level of irreproducibility \cite{shamir20} in a training model prior to its deployment, saving resources. In large scale, we also observed that it is sufficient to measure PDs between only two models.  This is
very suitable for cutting costs in such scale.

\subsection{Our Contributions}
Our main contributions are:
\bi
 \item We expose how critical irreproducibility can be to real world large scale systems, specifically recommendation systems, both to a user as well as to the development engineer.  As such, it is important to study and understand this problem theoretically and empirically, as well as develop methods that are able to mitigate it to reasonable levels.
 \item We demonstrate how smooth activations can be useful to mitigate irreproducibility in large scale production systems.
 \item We propose using \emph{Smooth reLU (SmeLU)} as a smooth activation to mitigate irreproducibility.  SmeLU has a much simpler mathematical form than other smooth activations.  This can lead to cheaper implementations, yet it gives trade-offs on reproducibility-accuracy comparable to or even favorable over much more mathematically complex smooth activations.
 \item We report results from a range of production recommendation systems (including content and application recommendation, and CTR prediction) in which we were able to deploy SmeLU in production with both reproducibility and accuracy gains.  This led to substantial simplifications, as well as training speed improvements of these systems, cutting down resource usage, and development times.
 \item We report results in which we were able to productionalize smooth activations from the wider family of \emph{generalized SmeLU (gSmeLU)} activations (that includes SmeLU) with accuracy gains over other activations, specifically in a distillation \cite{hinton15} pipeline.
\ei

\subsection{Related Work and Productionalization}
{\bf Reproducibility:}
Many factors contribute to irreproducibility in deep models \cite{damour20,fort2020deep,frankle2020linear,shallue18,snapp2021,summers21,zhuang2021randomness}.  The highly non-convex objective \cite{fort2020deep}, combined with nondterminism in training \cite{summers21} and underspecificaiton \cite{damour20} of over-parameterized deep networks, can lead training models to optima at different locations in a manifold or sets of optima.  Nondeterminism can emerge from the highly parallelized, highly distributed training pipelines, quantization errors, hardware types \cite{zhuang2021randomness} and more.  Slight deviations early in training due to these can lead to very different models \cite{achille17}).  A fixed initialization does not mitigate the problem.  Neither do standard methods like regularization, learning rates, or other hyper-parameter tuning, dropout, and data augmentation.  Among the techniques, some may reduce PDs.  However, that comes at the expense of degraded validation accuracy.  Degraded accuracy makes these methods undesirable in production systems. Warm starting models to solutions of previously trained models may not be a good idea, because without any guidance, a solution that leads to unfavorable metrics may be the one we anchor to.  Moreover, in a production development cycle, new models, architectures, and algorithms are being developed.  A new model may not always be aligned with a previous generation, and it may not be possible to warm start its now different parameters to those of a previous model.  With huge datasets in practical systems, enforcing determinism \citep{nagarajan18} in training is also not an option.  

A natural approach to improve reproducibility is using ensembles \cite{dietterich00}, and specifically {\em self} ensembles \cite{allen2020towards}, where multiple duplicates of the model are trained, and the final prediction is the average prediction of the duplicate components.  Ensembles have also been proposed and studied in the context of model prediction uncertainty (see e.g., \cite{lakshminarayanan17} and references therein).
Each component converges to a different optimum, and the average reduces the prediction variance.  Ensembles also reduce PD levels, and with different initialization of the parameters of each component, PD of the overall ensemble reduces even more.  However, in real world scale practical recommendation systems this can come at the expense of other costs. Unlike images, recommendation systems implement sparse machine learning problems, as we discuss in Section~\ref{sec:sys}.  A huge set of input features are learned, but every training example consists only of an insignificant fraction of the feature set.  Those features are mapped to embedding vectors that are concatenated as the input to the deep network.  Embeddings constitute the dominating fraction of model parameters.  Moreover, many models we consider train online with single visit per example, as opposed to batch learning with multiple visits of each example.  In this production regime, ensembles are, in fact, inferior in their prediction accuracy to single networks with the same number of parameters or with the same complexity.  For keeping equal complexity, a network component of an ensemble will have narrower embedding vectors and hidden layers than a single network counterpart.  Thus in a production system with limited parameter complexity, using ensembles may improve reproducbility, but at the expense of accuracy (which directly effects important downstream metrics). This is unlike image models \cite{kondratyuk2020ensembling,lobacheva2020power,wang2021wisdom} where the model learned parameters are dominated by internal link weights and biases.  Ensembles may make production models more complex and harder to maintain.  Production models may include various parts with special handling of various special cases.  Interactions of such parts with multiple duplicates as well as with their ensemble can build up to a substantial technical debt (see, e.g., \cite{sculley14}).  Finally, as we have observed in real cases, having multiple sets of the same sparse input features, as required with ensembles, creates a bottleneck that can slow training substantially.

Compression of deep networks into smaller networks that attempt to describe the same information is the emerging area of \emph{distillation} \citep{hinton15}.
Predictions of a strong \emph{teacher} train a weaker \emph{student} model.  The student is then deployed.  This approach is very common if
there are ample training resources, but deployment is limited, as for mobile network devices.  
\emph{Co-distillation} \cite{anil18} (see also \cite{zhang18}) embraces training ensembles and distillation to address irreproducibility.  Instead of unidirectional transfer of knowledge, several models in an ensemble distill information between each other, attempting to agree on a solution.  The method requires more training resources to co-train models, but deployment only requires a single model (which can be an ensemble by itself).  The deployed model follows the PD levels of the training ensemble, but may exhibit some degradation in accuracy due to forcing the components to agree on a solution they may not prefer.

A somewhat opposite approach; \emph{Anti-Distillation} was proposed in \cite{shamir20}, again, embracing ensembles, with an
additional (decorrelation) loss that forces their components away from one another.  Each component is forced to capture a (more) different part of the objective space, and as a whole, the predictions of the ensemble are more reproducible.  This may still come at the expense of accuracy.  However, we were able to deploy anti-distillation models with reproducibility improvements and without any accuracy loss.  In such models, different components, such as linear components of the model, were able to compensate for the accuracy loss caused by the anti-distillation decorrelation loss.

Distillation from the same teacher can reduce PDs.  However, like warm starting to the same solution, it may not necessarily be a desired solution, as it may anchor predictions to an unfavorable teacher.  Unlike warm starting, distillation does not need to align parameters to those of the teacher.  Using ensemble teachers trained to be reproducible (potentially with anti-distillation) \cite{shamir20ed} with a more accurate student and/or teacher can improve reproducibility of a deployed model forcing it towards a more reproducible solution, which may not be just that of a disfavored teacher.  This method can give a deployed model with the accuracy of a single component model, and reproducibility of the ensemble teacher.  An ensemble does not need to be deployed, but must be trained and maintained as a distillation teacher model.
Other recent approaches to address the irreproducibility problem attempted to anchor the a solution to some constraint \cite{bho21, shamir18} but also degrade performance by constraining the model's ability to converge to a better solution.

{\bf Activations:}
The \emph{Rectified Linear Unit (ReLU)} activation \cite{nair10} has been instrumental to deep networks in recent years.
With back-propagation it gives simple updates, accompanied with superior accuracy.
Due to its non-smoothness, ReLU imposes an extremely non-convex objective surface.  With such surface, the order in which updates are applied is a dominant factor in determining the optimization trajectory, providing a recipe for irreproducibility.

Various recent works started challenging the dominance of ReLU, exploring alternatives.
Overviews of various activations were reported in \cite{nwankpa18, pedamonti18}.  Variations on ReLU were studied in \cite{jin15}.
Activations like SoftPlus \cite{zheng15}, \emph{Exponential Linear Unit (ELU)} \cite{clevert15},
\emph{Scaled Exponential Linear Unit (SELU)} \cite{klambauer17, sakketou19, wang17}, or \emph{Continuously differentiable Exponential Linear Unit (CELU)}
\cite{barron17} were proposed, as well as the
\emph{Gaussian Error Linear Unit (GELU)} \cite{hendrycks16}.
The \emph{Swish} activation \cite{ramachandran17} (approximating GELU)
was found through automated search to achieve accuracy superior to ReLU.
Additional GELU like activations; \emph{Mish} \citep{misra19},
and \emph{TanhExp} \citep{liu20tanhexp}, were recently proposed.
Unlike ReLU, many of these activations are \emph{smooth} with continuous gradients. Recent work \cite{bresler20} smoothed ReLU taking only initial coefficients of its Fourier series. 
Studies \cite{mhaskar97} (see also \cite{du19, lokhande20}) suggested potential advantages to smooth activations, where recent work \cite{xie20}, inspired by our results, applied smooth activations to adversarial training.


\section{System Overview}
\label{sec:sys}
We consider a supervised learning problem in recommendation systems.  Logs of past \emph{recommendation sets\/} constitute the training dataset.  A recommendation set can be 
a stream of past recommendations that a user engaged with, or it can be an outcome set provided for a specific query issued by the user.  
A recommended item is assigned a label based on the user response or engagement with the item.  In the simplest binary case, which we focus on, the label is either positive, i.e., the user engaged with the item, or negative, the user did not engage with the item.

Many such systems are in the sparse regime.  There is a huge selection of items the user can engage with, and a huge set of features that can describe the request as well as the items themselves.  Features can be properties of the request, the item, the combination of the two, the user, the recommendation user interface, the recommendation rendering, and more.  A single recommendation item, represented as a labeled example in the training dataset, will have only an insignificant portion of features present.  Using standard notation, the $t$ example is the pair $\{\xvec_t, y_t\}$, with $\xvec_t$ being a feature value vector and $y_t \in \{0,1\}$ the label.  In such sparse problems, most entries in $\xvec_t$ are 0. Some features are binary; either the $i$th feature is present, with the respective component $x_t^i$ of $\xvec_t$ taking value $1$, or absent, with $x_t^i = 0$. Some features are categorical, i.e., from a group of features, only one can be present, although for some categories, multiple ones can be in a given example.

Models for such systems can train on the training dataset with multiple \emph{epochs}, where in each epoch, all or a subset of the currently available training dataset is (re)visited.  A trained model is then deployed to provide recommendations to live user requests.  These recommendations are ordered by engagement rates predicted by the model, or joined with other signals, such as bids in CTR, and shown to the user based on combination of the predicted engagement rates and the other signals.  The requests, their outcomes, and users' engagements with the outcomes produce more training examples that can be fed into the training dataset, and used to refresh models.  Models can be retrained from scratch on refreshed datasets, or trained continuously when additional data is available.

For some recommendation systems, such as device application installation recommendations, except in special cases, current world trends may have smaller effects on engagement rates.  However, in others, such trends can affect which content a user is engaged with, or which ads users tend to click on.  Trend as well as other temporal nonstationarities justify \emph{online} learning, where models train with a single visit to each data point, and examples are ordered in a roughly chronological order.  With ample training data, which many such systems have, this reduces overfitting to the training data.  (Some of the real systems we consider train online, although, others, for which the time is not as critical, may use data shuffling with different schedules.)  For efficiency, production systems have to be mini-batched, parallelized and distributed.  Thus even with online systems, training is not per-example online, and model update ordering is only roughly chronological.  Training accuracy (and other training metrics) can (still) be evaluated with \emph{progressive validation} \cite{blum99}, where prediction on each example is evaluated relative to the true label before the model trains on the example.  We focus the exposition on online learning production systems, but the methods described also apply to batch training systems.  Specifically, some results reported in Section~\ref{sec:exp} did not use an online schedule.

Recommender deep neural networks can be fully connected or of other architectures.  Due to the extreme sparsity, models can train mainly on individual item engagement label rates, although other losses can also be introduced for various purposes.  Numerical input features act as inputs to the network.  Categorical features are mapped into embedding vectors, each vector representing a category.  All features/embeddings are concatenated into an input layer of the deep network.  The embeddings constitute the majority of the model parameters, but as described, only an insignificant portion of the stored embedding vectors is present in any training example.  Embeddings are stored in highly distributed systems, and the correct vector must be fetched for any training example.  Hence, fetching embeddings can be costly.  This cost scales with ensembles that store multiple copies of embedding vectors for the same feature, all of which must be fetched for a given example.  This can substantially slow ensemble training (unless model components are shared).

Taking the embedding layer as inputs, a network of several hidden layers is used to produce an engagement rate output.  Since the sparse parameter space is dominated (by far) by embeddings, deep networks can be rather shallow with single digit layer count. Let $\avec^\ell$ denote the activation output of layer $\ell$.  The embeddings constitute layer $\ell=0$. Starting with layer $\ell=1$, a nonlinearity $f(\cdot)$ is applied on the output of the previous layer.  Then, the layer produces an output $\avec^\ell = W^\ell \cdot  f(\avec^{\ell-1}) + \bvec^{\ell}$ where $W^\ell$ and $\bvec^{\ell}$ are link weight matrices and bias vectors, respectively, that are learned by the network.    For the output layer, $\ell=L$, $W^L$ can have a single row (although if multiple losses or outputs are needed, it can have more rows).  For binary labeled models, we use logistic regression with logarithmic cross entropy loss.  The output of layer $L$ is converted to probability of positive engagement with the logistic (Sigmoid) function $\sigma ( x ) \dfn 1 / (1 + \exp(-x))$.  The positive engagement probability predicted by the model is $\hat{y}_t = \sigma(a^L)$.  It is compared against the true label $y_t$ producing loss whose gradients propagate to the network. Some form of Stochastic Gradient Descent (SGD) can be used to train the models, including the embeddings, and the weights and biases.  Because of the problem sparsity, per-coordinate learning schedules as those in AdaGrad \cite{duchi11} are used, especially for training the embeddings.  Actual production models may be a lot more complex with additional architecture, factorization, constraints and components, as well as use more complex optimizers.  However, such complications are beyond the scope of this paper, and do not change the results and conclusions.

Hidden layer activations are clipped within some range to ensure numerical stability.  Together with capping, some form of normalization should be applied to limit the range of the signal.  Without such normalization, an extremely non-smooth objective surface is attained, which at the very least encourages model irreproducbility.  Weight normalization \citep{salimans16}, layer normalization \citep{ba16}, or batch normalization \citep{ioffe15} can be applied.  For shallow networks with sparse embedding inputs, batch normalization may not be ideal.  Layer norm and a slightly different form of weight normalization have shown beneficial to our production systems.  With this form of weight norm, an $L_p$ norm of a row of $W^{\ell}$, constituting the matrix weights incoming to some output neuron of the layer, is kept at a fixed value $v$.  Specifically, with $p=2$, let $\wvec_j^\ell$ denote the $j$th row of $W^\ell$.  Then, $\wvec_j^\ell$ is replaced by $\tilde{\wvec}_j^\ell \dfn v \cdot \wvec_j^\ell / \| \wvec_j^\ell \|_2$.  One advantage of this weight norm is that in deployment, normalization can be computed once for all weights, whereas with layer norm, normalization must be computed on the activations for every inference.

A very important trait of mature large scale production recommendation systems is that they have been highly optimized. From our experience, even very small accuracy improvements of small fractions of a percent of the training loss can make a huge difference to the application system's overall performance. For binary-labeled models, in addition to training loss, ranking loss, commonly computed as Area Under the Curve (AUC) loss is also important.
\be
\label{eq:auc_loss}
 L_{\text{AUC}} \dfn \frac{1}{N} \cdot \sum_{t:y_t = 1} \cdot \sum_{\tau:y_\tau = 0}
 I(\hat{y}_\tau > \hat{y}_t),
\ee
where $I(\cdot)$ is the indicator function, and $N$ is the total number of positive-negative label pairs in the validation corpus.  Such a loss can also be computed per query (or request) (PQAUC). Training and ranking accuracy tend to be well correlated with many of the practical system's objectives. Therefore, resources can be saved by eliminating models with bad accuracy metrics without need for live deployment.

\bef
 \centerline{
 \includegraphics[width=0.22\textwidth, clip=]{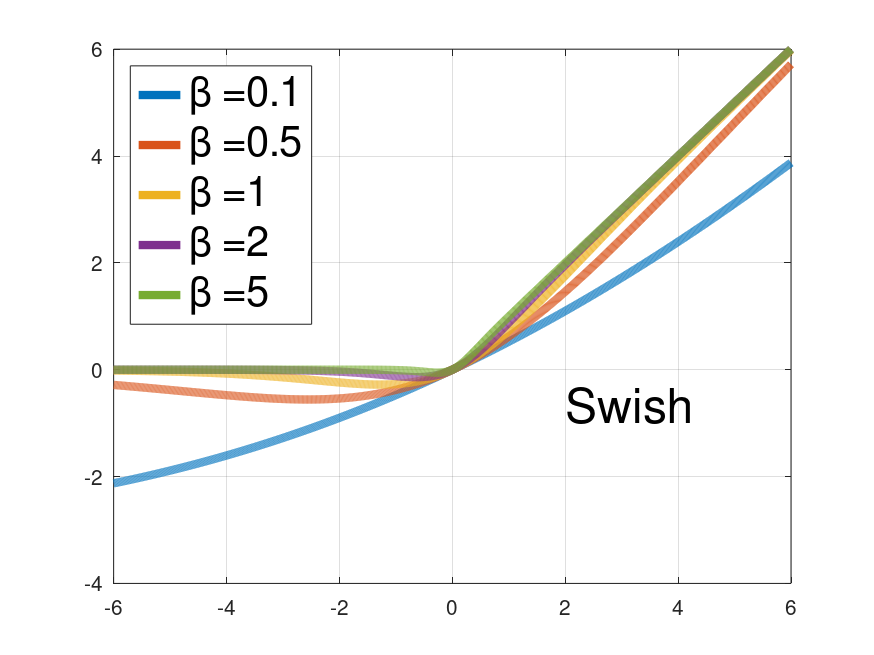}
 \includegraphics[width=0.22\textwidth, clip=]{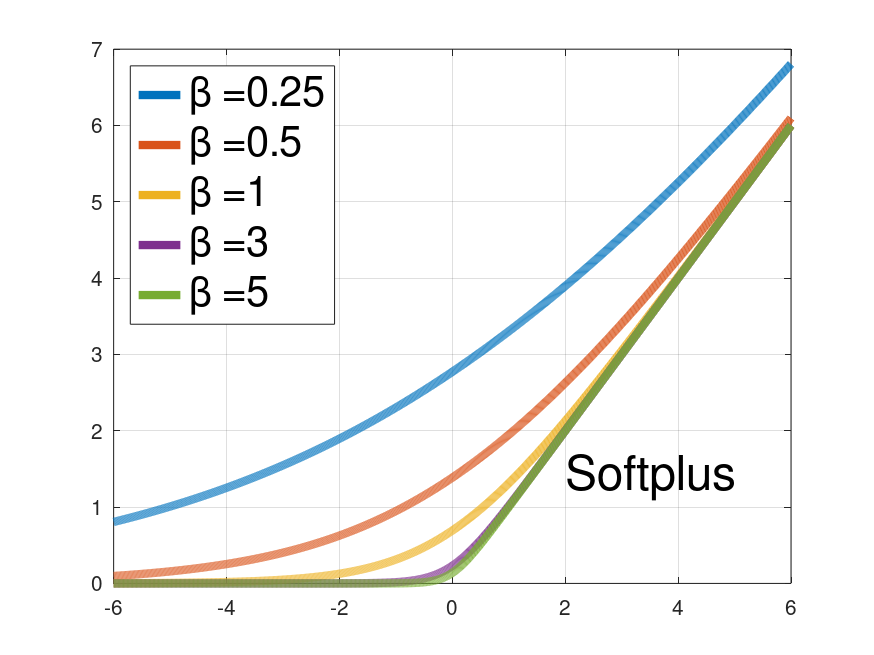}
 \includegraphics[width=0.22\textwidth, clip=]{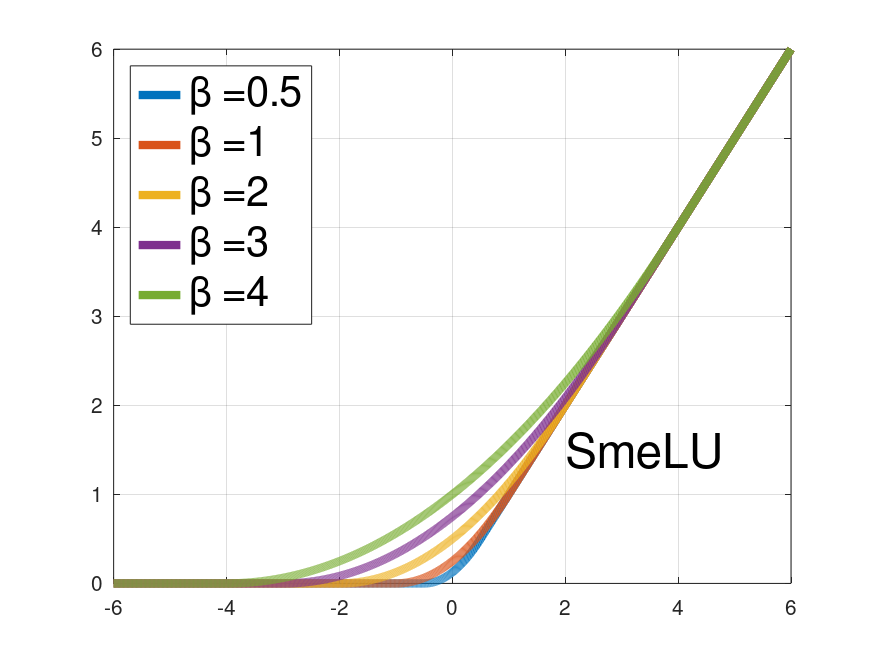}
  \includegraphics[width=0.22\textwidth, clip=]{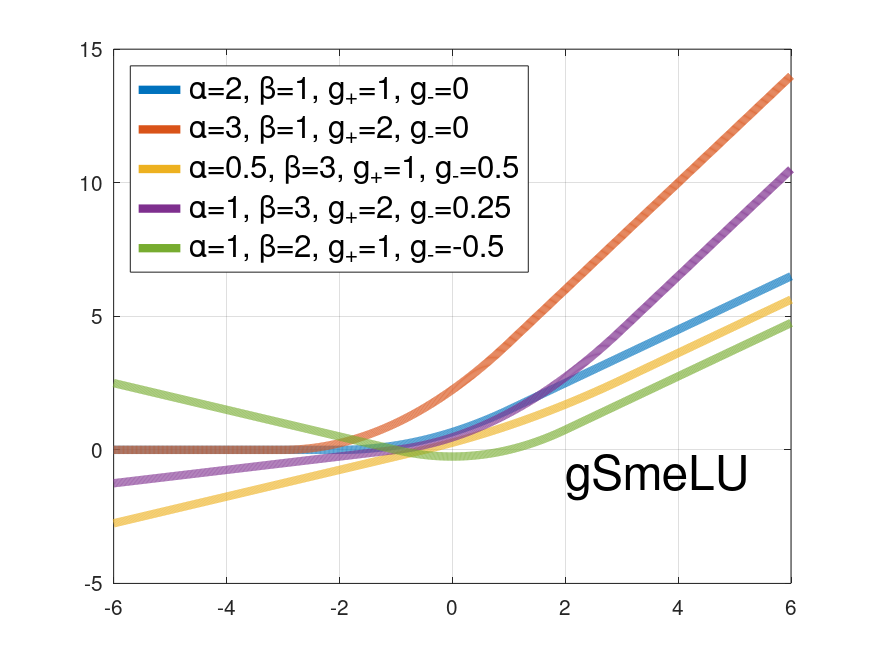}}
 \centerline{
 \includegraphics[width=0.22\textwidth, clip=]{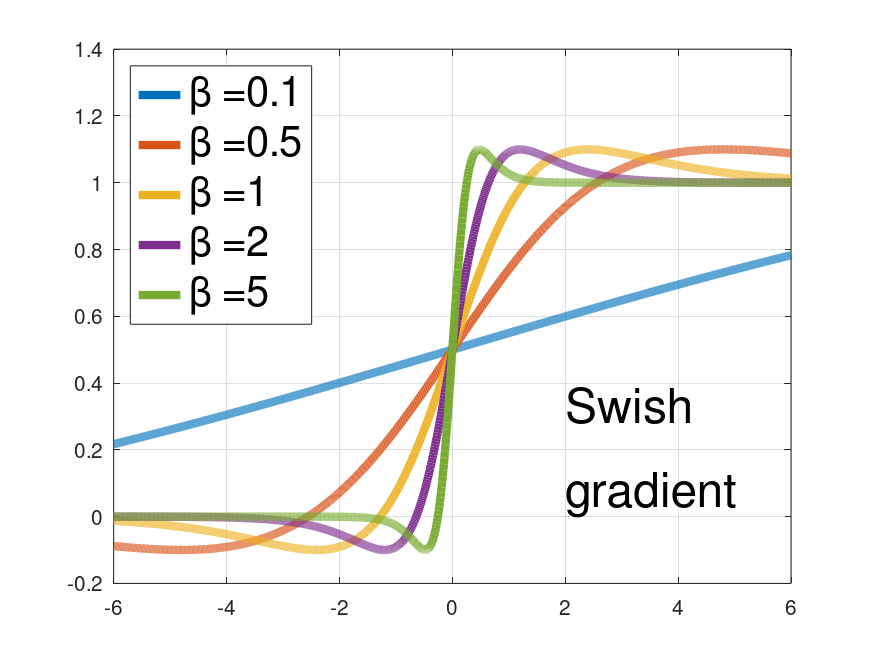}
 \includegraphics[width=0.22\textwidth, clip=]{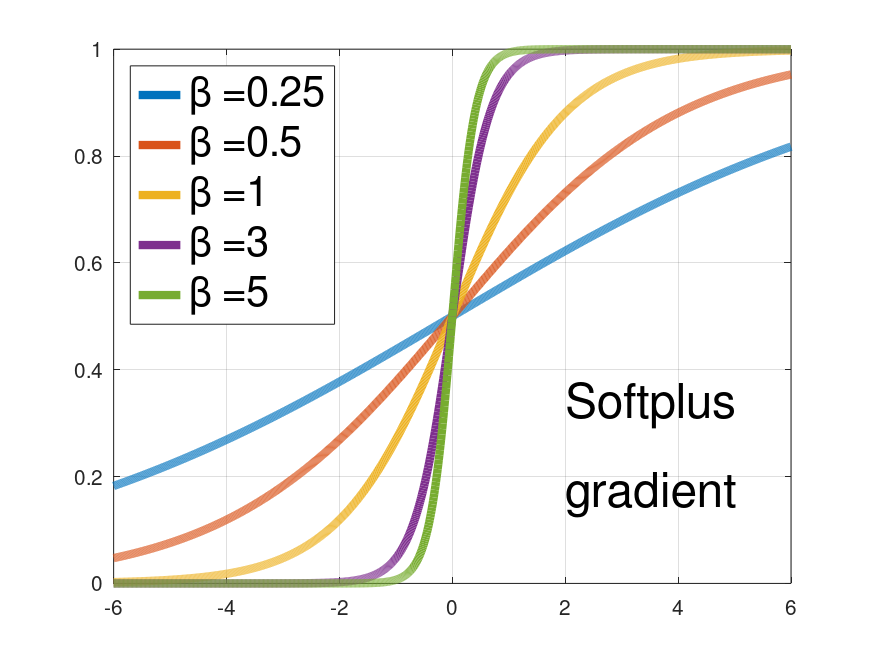}
 \includegraphics[width=0.22\textwidth, clip=]{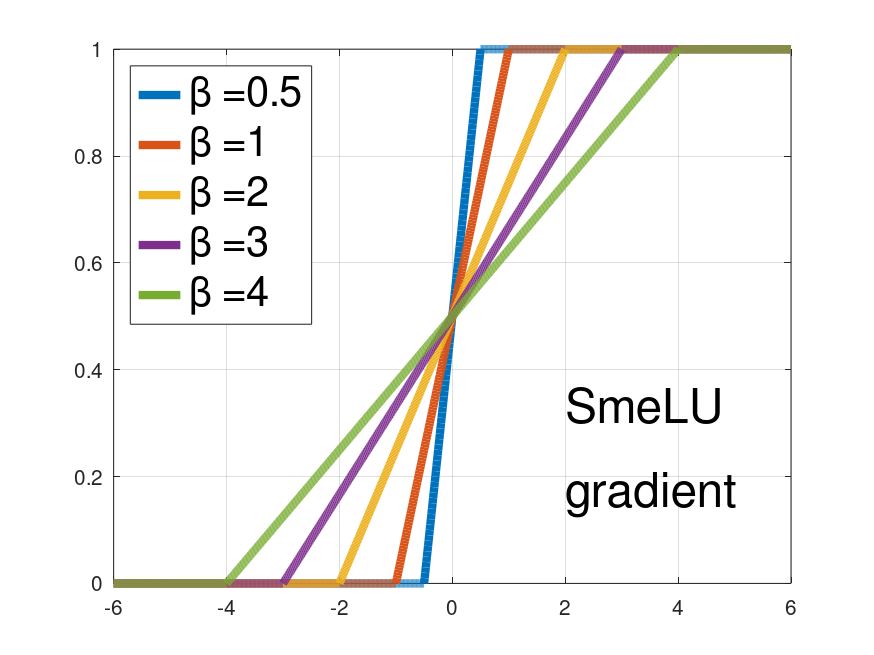}
  \includegraphics[width=0.22\textwidth, clip=]{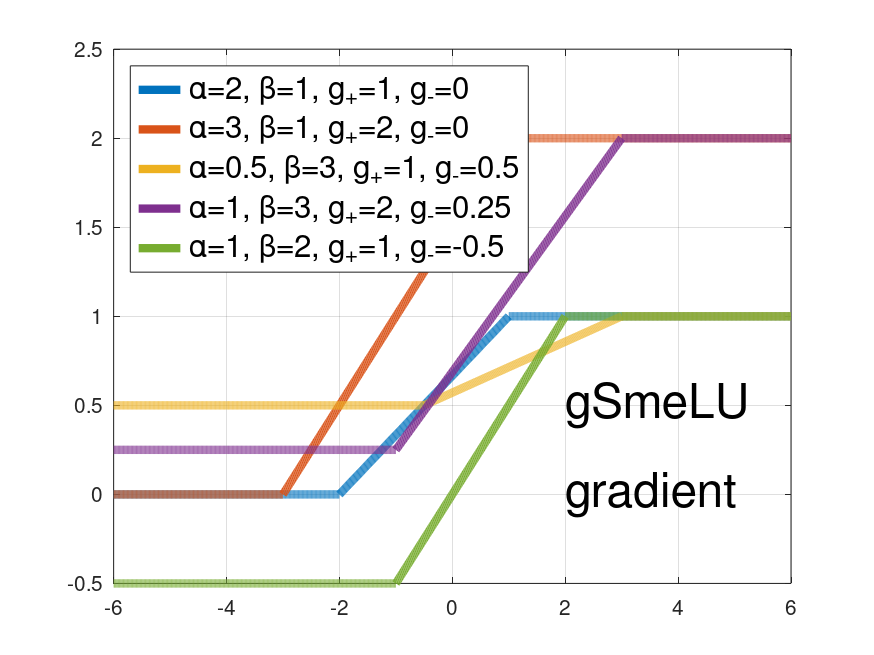}}
 \vspace{-.2cm}
 \caption{Smooth activations (top) and their gradients (bottom) for different $\beta$ parameter values.}
\label{fig:smooth}
\enf


Deployed production models may be constrained in complexity, amount of training data, and other system considerations. As such, they may not be able to produce metrics that can be produced without these limitations.  Unconstrained \emph{teacher} models can be trained and used with distillation \cite{hinton15} to transfer such gains to the \emph{student} to be deployed.  Distillation can be advantageous for reproducibility beyond the methods described in Section~\ref{sec:introduction}.  Some techniques trade off reproducibility to accuracy.  We can train more accurate models, with degraded reproducibility, and use distillation to distill the accuracy of these models to more reproducible student models. This approach has been used with gSmeLU to improve production model performance, as we report in Section~\ref{sec:exp}.

\section{Prediction Difference}
\label{sec:pd}
Engagement rate \emph{Prediction Difference (PD)} for individual examples is correlated with system level reproducibility performance metrics.  For content recommendations, such differences can lead to different or swapped recommendation sets.  
When predictions are used for auctions, PDs affect the auction outcomes.  Unlike classification applications that focus on the final label, in recommendation systems, the exact predicted engagement probabilities can make a difference.  Theoretically, we can use different statistics averaged over multiple models \cite{chen20,shamir20,yu2021dropout} such as standard deviations or KL divergences.  In \cite{shamir20}, various $L_p$ norms relative to an average prediction of a set of models were considered.  However, training multiple models is expensive in large scale production systems. Thus a metric with minimal training costs is desirable.  PDs between two models in our large scale systems tend to settle to roughly the same values for different pairs of the same model.  Thus it is (usually) sufficient to train one pair of models and measure PD between them.  We also want a metric that is not sensitive to actual predicted engagement rates. We define the \emph{relative PD\/} for example $t$ as the absolute difference between predictions $\hat{y}_{t,1}$ and $\hat{y}_{t,2}$ of models $1$ and $2$ normalized by their average prediction.  Then, the model relative PD, averaged on all examples, is given by
\be 
 \Delta_r \stackrel{\triangle}{=} \frac{1}{T} \cdot \sum_t \frac{2 \cdot |\hat{y}_{t,1} - \hat{y}_{t,2}|}{\hat{y}_{t,1} + \hat{y}_{t,2}}.
\ee

\section{Smooth Activations}
\label{sec:smooth}
Several forms of smooth activations have been recently proposed, all attempting to preserve ReLU's shape with a smooth form. Many can be parameterized with a single parameter, denoted as $\beta$. Let $\text{erf} (\cdot)$ be the standard error function and $\Phi(\cdot)$ the standard normal CDF.  Functional forms of some smooth activations are:
\bea
 y_{\text{SoftPlus}} &=& \frac{1}{\beta} \cdot \log \left [ 1 + \exp \left ( \beta \cdot x \right ) \right ]  
 \label{eq:softplus}
 \\
 y_{\text{Swish}} &=& x \cdot \sigma  \left ( \beta \cdot x \right )
 \label{eq:swish}
 \\
 y_{\text{GELU}} &=& x \cdot \frac{1}{2} \cdot \left [ 1 + \text{erf} \left ( \beta \cdot x / \sqrt{2} \right ) \right ] = x \cdot \Phi (\beta \cdot x)
 \label{eq:gelu}
\eea
Activations such as Mish and TanhExp are similar to GELU and Swish, and did not add benefit to accuracy-reproducibilty trade-offs in our systems.  Others, such as SELU did not perform as well. GELU can be approximated by Swish with $\text{GELU}_\beta (x) \approx x \sigma(\sqrt{8/\pi} \beta x)$.
Fig.~\ref{fig:smooth} shows the different activations and their gradients for different values of $\beta$.  The parameter $\beta$ controls the width of a transition region from approaching $0$ on the left towards a slope $1$ on the right.  For activations in \eref{eq:softplus}-\eref{eq:gelu}, a smaller $\beta$ gives a wider region, and a larger $\beta$, a narrower one.  When $\beta \rightarrow \infty$, the activation resembles ReLU, and when $\beta\rightarrow 0$, it is close to linear. 
Activations like GELU and Swish are not monotonic, and have a region in which they change the direction of the gradients. Also, they do not have a stop region, i.e, strictly $0$, and a clear slope $1$ region. Activations, such as GELU  must be either numerically computed or approximated by other functions. Such activations require more complex hardware implementation, especially with cheaper, simplified hardware that supports only a limited number of operations. These properties can make deployment error-prone, expensive, or slow. 

\bef
\centerline{
 \includegraphics[width=0.22\textwidth, clip=]{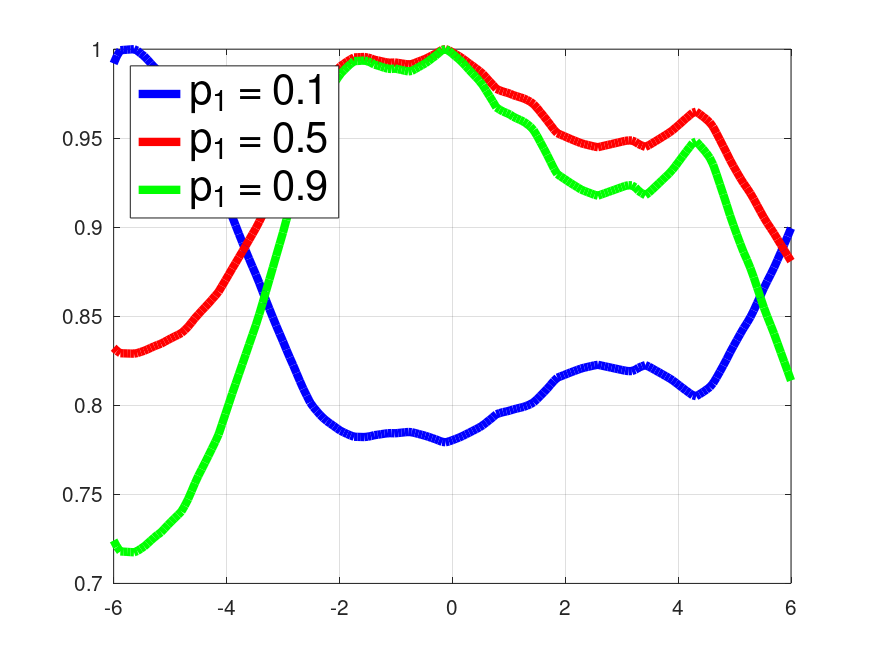}
 \includegraphics[width=0.22\textwidth, clip=]{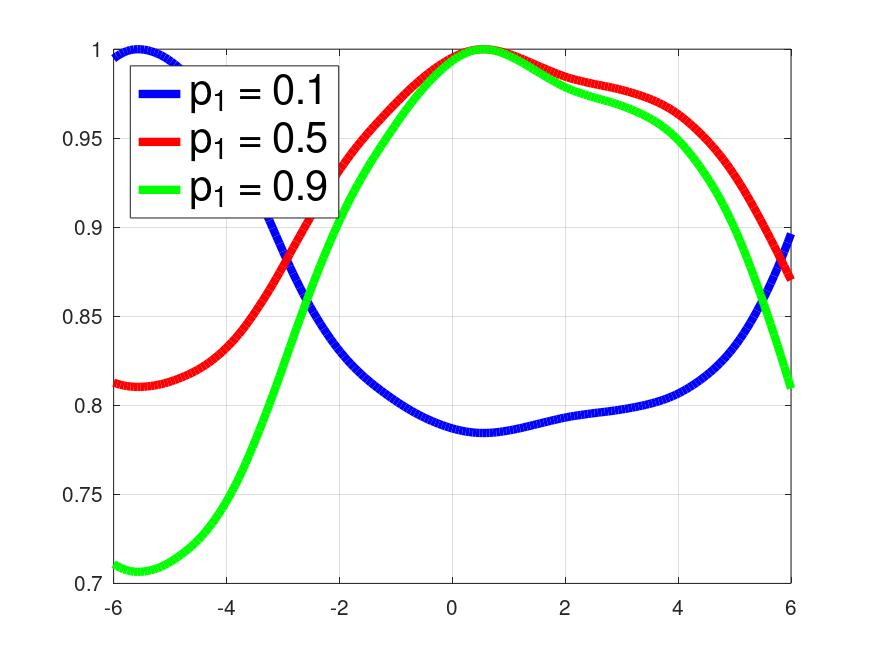}
 \includegraphics[width=0.22\textwidth, clip=]{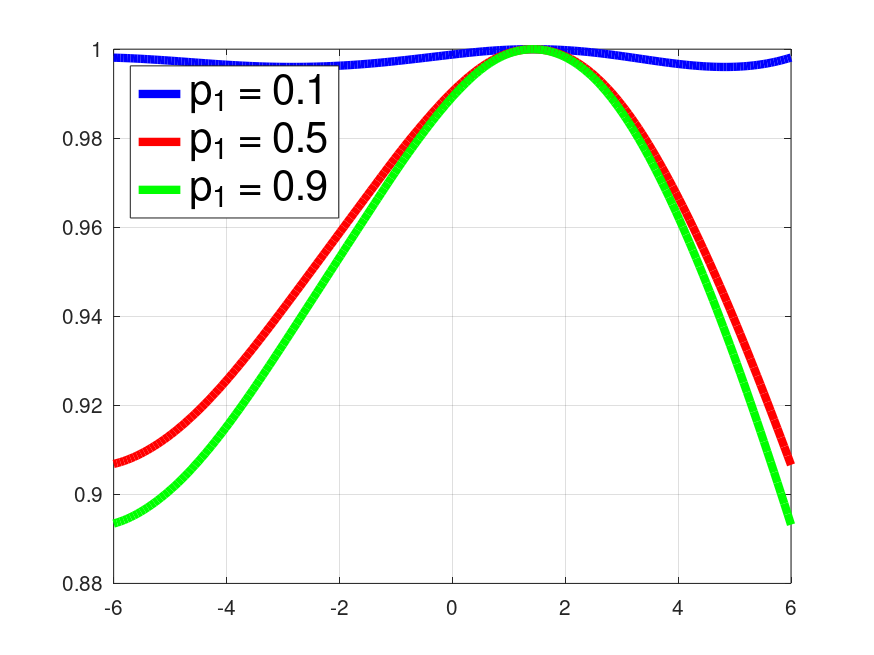}
 \includegraphics[width=0.22\textwidth, clip=]{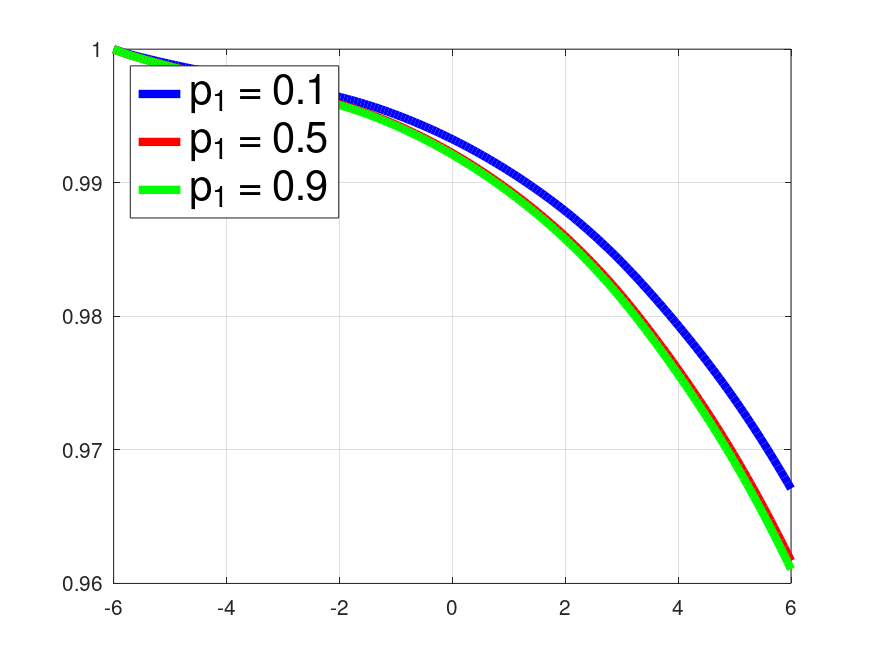}}
\centerline{
 \includegraphics[width=0.22\textwidth, clip=]{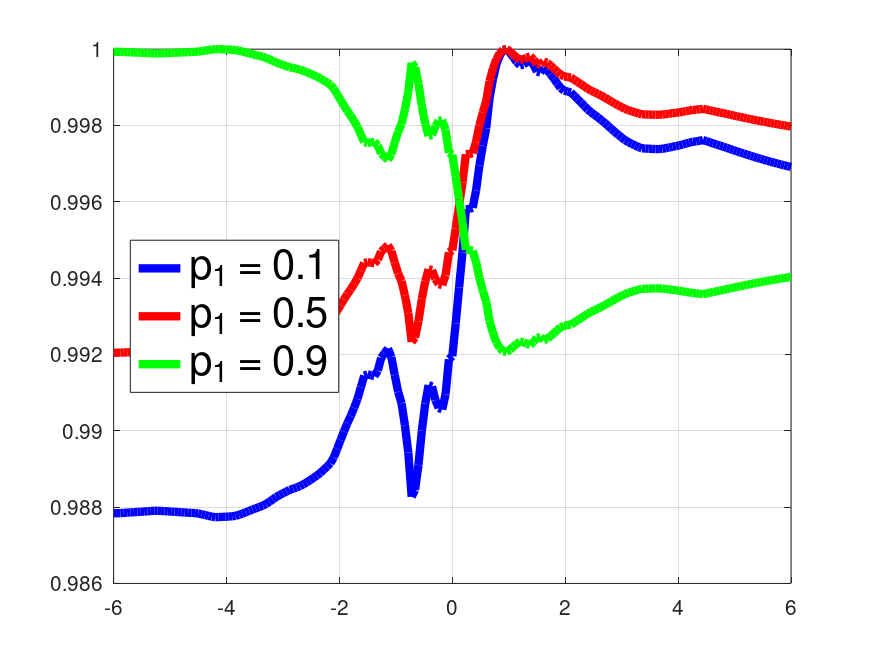}
 \includegraphics[width=0.22\textwidth, clip=]{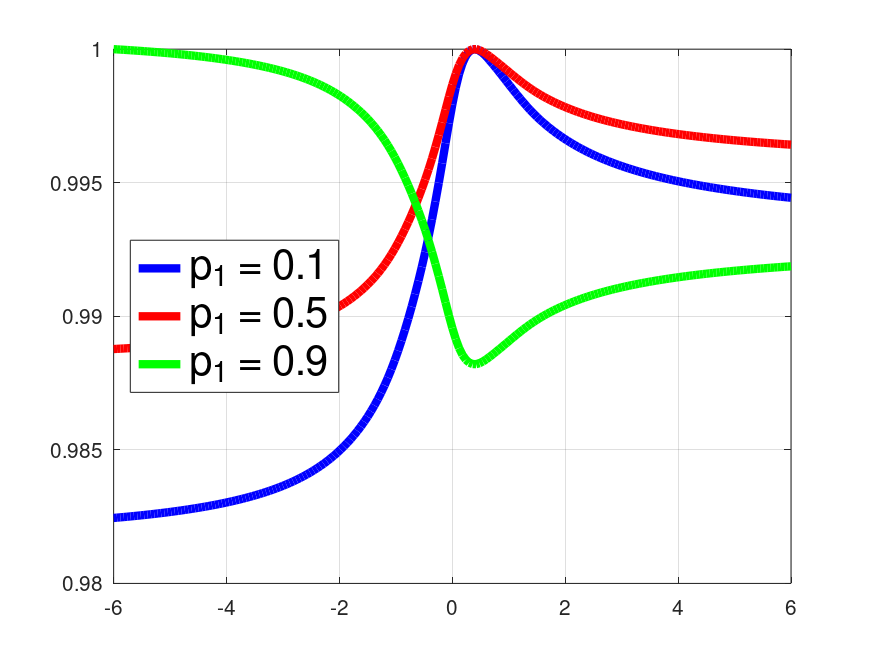}
 \includegraphics[width=0.22\textwidth, clip=]{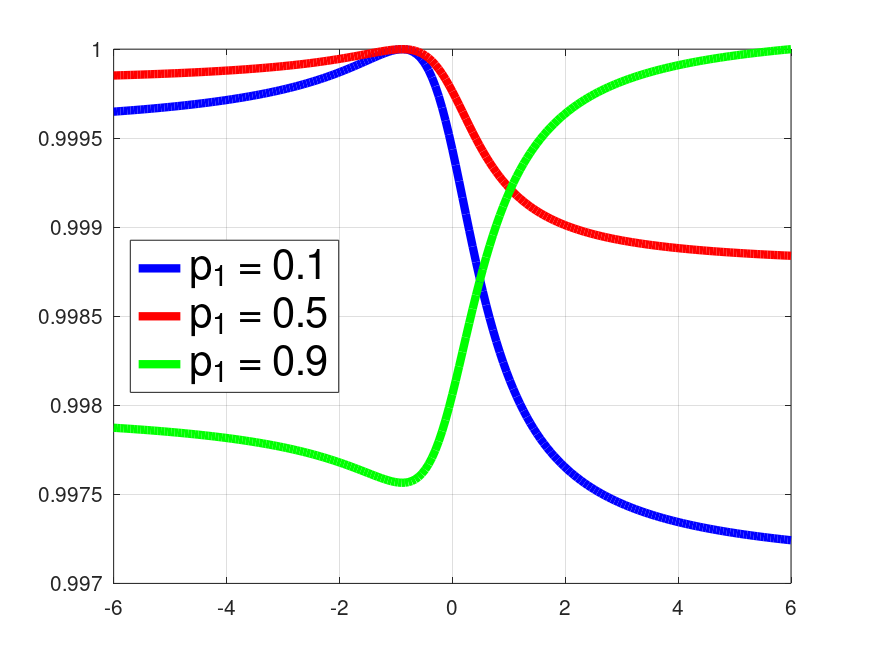}
 \includegraphics[width=0.22\textwidth, clip=]{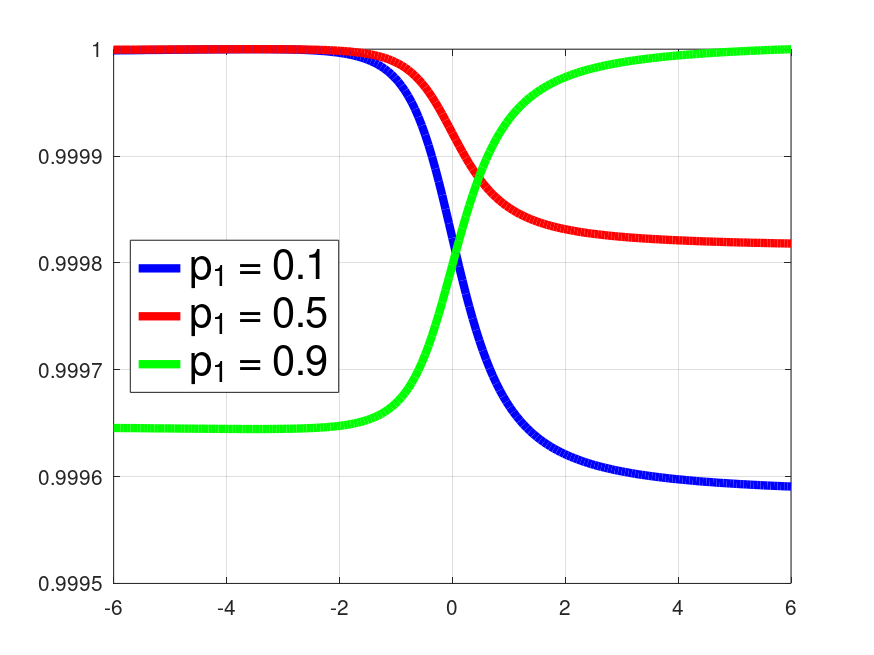}}
\centerline{
 \includegraphics[width=0.22\textwidth,height=0.05\textwidth, clip=]{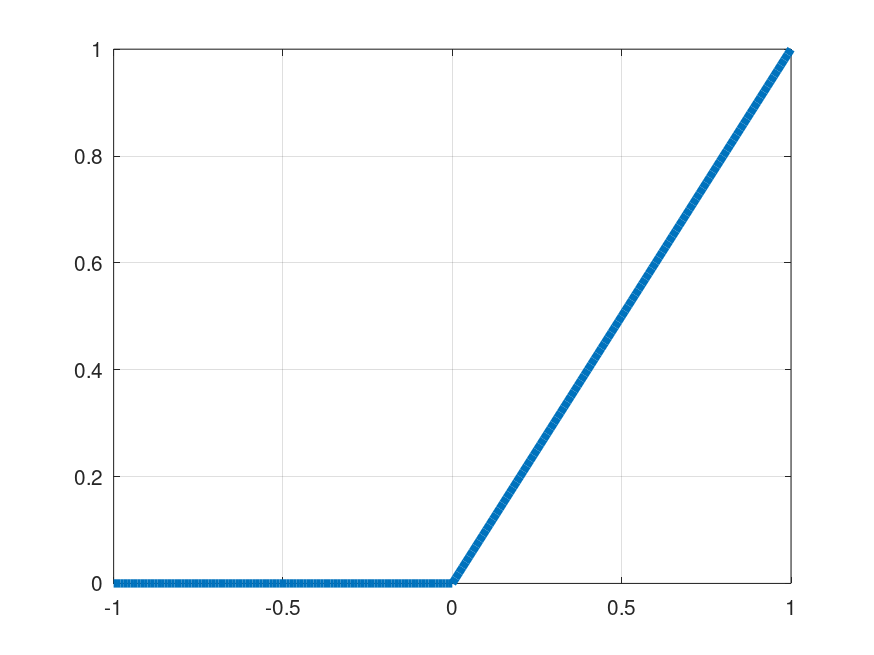}
 \includegraphics[width=0.22\textwidth,height=0.05\textwidth, clip=]{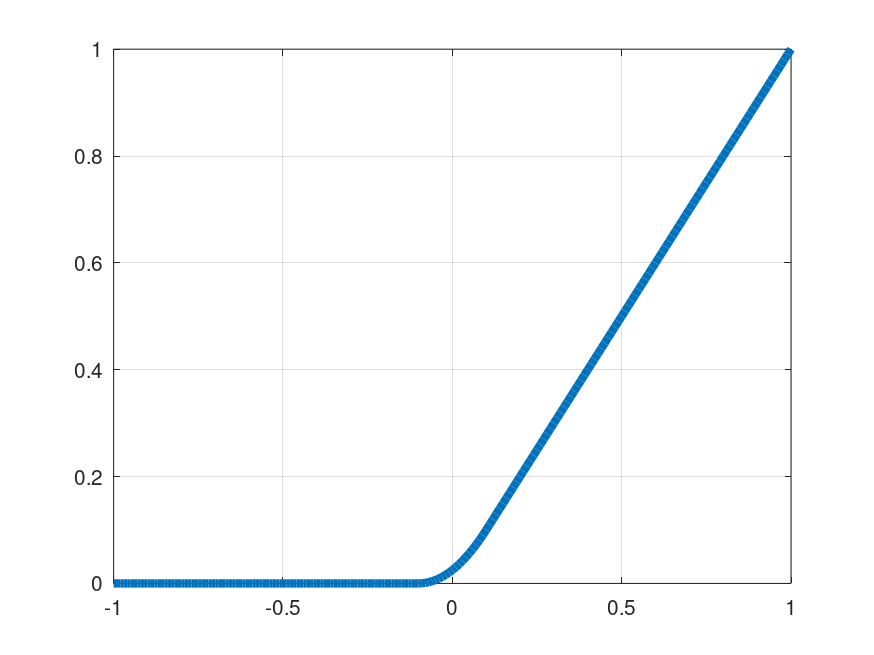}
 \includegraphics[width=0.22\textwidth,height=0.05\textwidth, clip=]{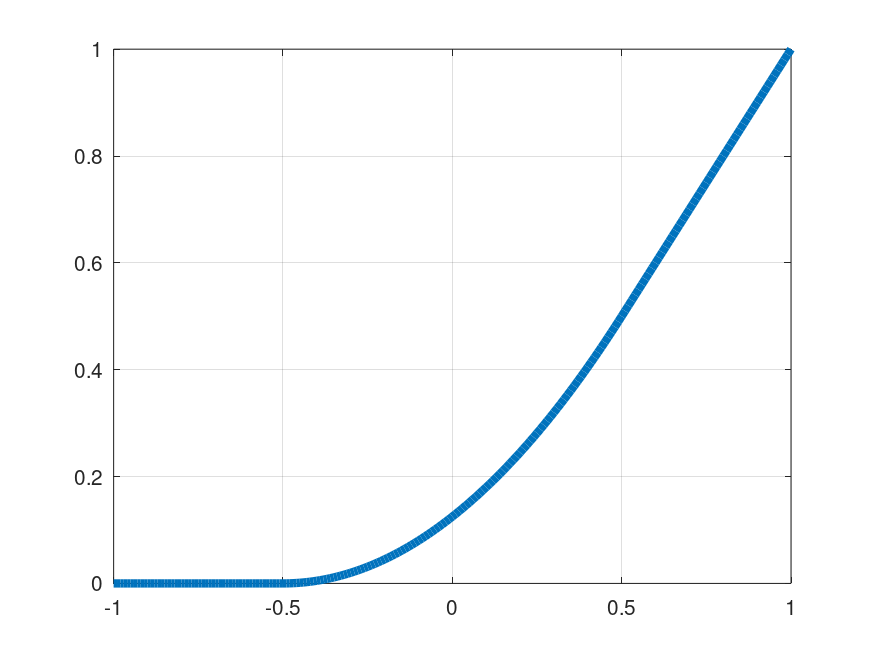}
 \includegraphics[width=0.22\textwidth,height=0.05\textwidth, clip=]{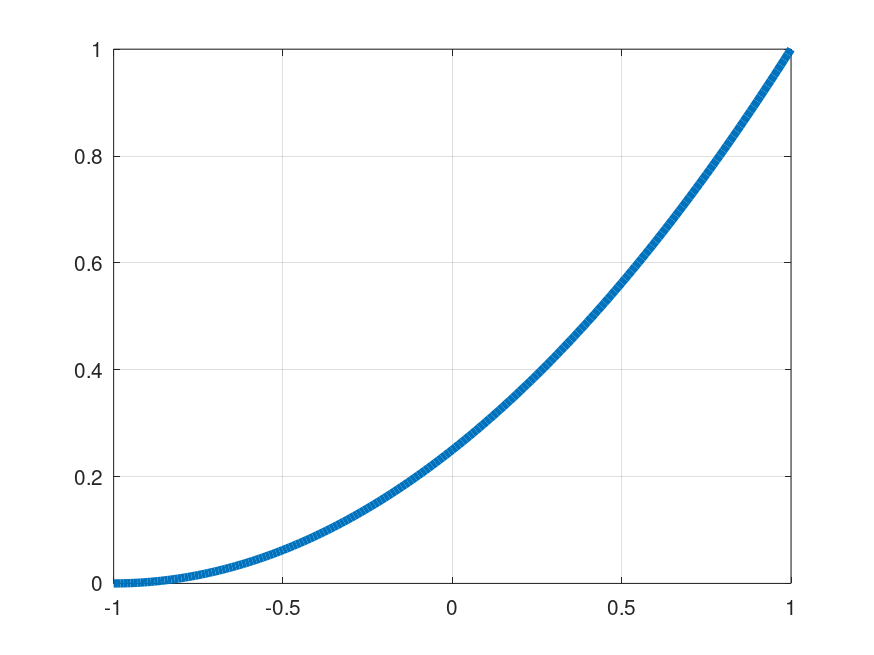}} 
 \caption{Normalized loss objective curves (scaled to fit the same axes) as function of a single input to a random network with $5$ hidden layers of dimensions $[256,128,64,32,16]$ activated by ReLU, and SmeLU with different $\beta$ values.  Activated values are clipped to $\{-6,6\}$.  The last layer outputs a scalar score $s$.  Then, for positive label probability $p_1$ (noted on the graphs), the graphs show logistic loss $L = p_1 \log(1 + \exp(-s)) + (1-p_1) \log (1 + \exp(s))$.  Top row: link weights are from a normal distribution $\mathcal{N}(0,25)$, biases from $\mathcal{N}(0,0.25)$, and weight normalization is applied.  Middle row: Link weights and biases drawn from $\mathcal{N}(0,1)$, and layer norm is applied.  Bottom row: the activation; ReLU on the left, and SmeLU with $\beta=0.1,0.5,1$ in the remaining columns.} 
\label{fig:1d_obj}
\enf

ReLU's noncontinuous gradient partitions the parameter space into regions with multiple local minima.  Some equal, but may provide different model predictions. Training randomness leads the model into a region, and converges to its local minimum.  Sudden jumps in gradient, like in ReLU,
may lock some hidden units at some states, requiring others
to compensate by moving towards their nearest optima.  If this happens early in training, it determines
the trajectory of the optimization.  Different examples and update orders thus lead to different trajectories.  Smoother activations give a model fewer opportunities to diverge. Wider transition regions allow gradients to change more slowly, and units in the
network to gradually adjust.  Wider transition regions are closer to more reproducible linear functions.
However, a good activation must also retain good accuracy.  Softplus, for example, may not achieve accuracy as good as other activations.  Roughly, reproducibility improves by widening activations' transition regions, but accuracy moves in the opposite direction.  However, as observed on our datasets, 
there exist tuning points where smooth activations outperform ReLU on both accuracy and reproducibility.  This is demonstrated in Section~\ref{sec:exp}. 
Smooth activations thus provide a tuning knob (the parameter $\beta$) to trade off between accuracy and reproducibility.  Trade-offs as we described are usually  dataset, model architecture, and model configuration dependent.

\section{Smooth reLU - SmeLU}
\label{sec:smelu}
As mentioned above, there are several disadvantages to activations in \eref{eq:softplus}-\eref{eq:gelu} and others alike. In this section, we introduce a new, simple smooth activation SmeLU, that gives competitive accuracy-reproducibility trade-offs.  We then generalize the methodology to describe a larger family of activations.

\bef
\centerline{
 \includegraphics[width=0.9\textwidth, clip=]{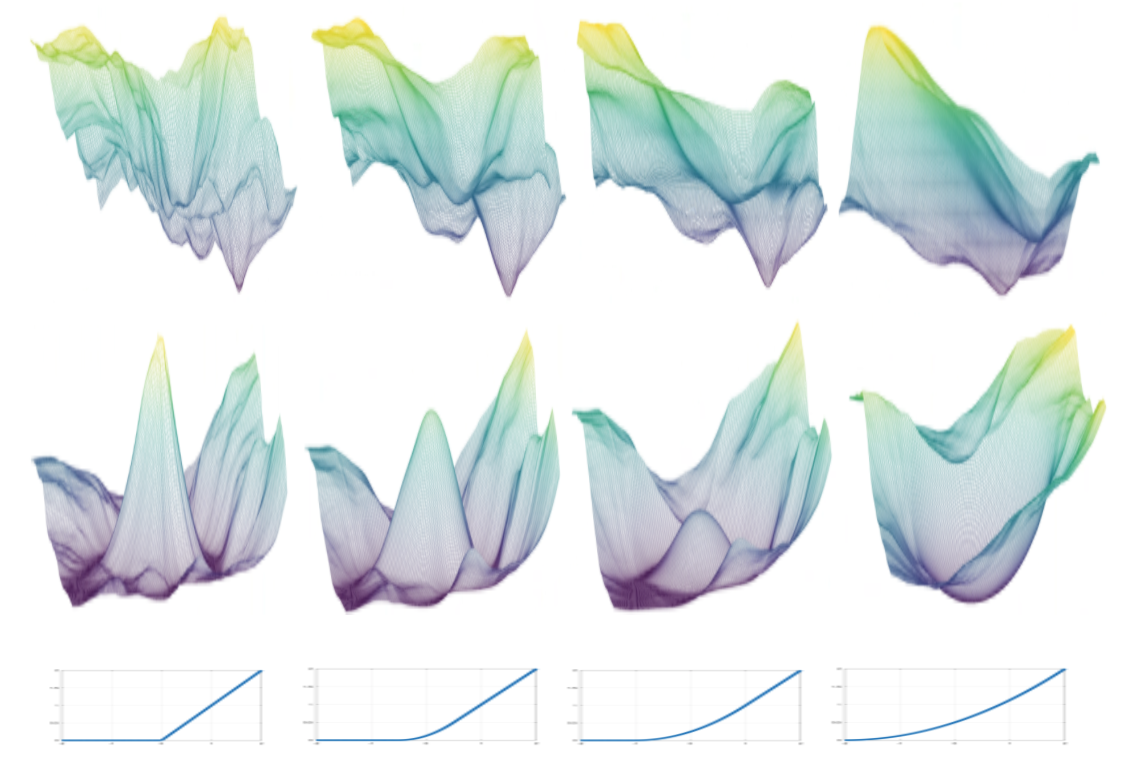}}
  \caption{Loss surfaces of networks as described in Fig.~\ref{fig:1d_obj} with two inputs, for ReLU and SmeLU with different $\beta$ values for two different losses.  Network weights and biases are drawn from $\mathcal{N}(0,1)$.  Weight normalization is applied. Top row: Logistic loss with positive probability $p_1=0.1$, and activations clipped to $\{-6,6\}$.  Middle row: Linear regression loss relative to $-2$, and  activations clipped to $\{-4,4\}$.  Bottom row: the activation; ReLU on the left, and SmeLU with $\beta=0.5,1,2$ in remaining columns.} 
\label{fig:3d_obj}
\enf

\subsection{SmeLU}
Using linear and quadratic pieces \cite{huber92} but asymmetrically (unlike \cite{huber92}), we design SmeLU to keep ReLU's functional form, including monotonicity, complete stop on the left, and gradient of 1 on the right. Multiple pieces are joined with smoothness, continuous gradient, constraints.  We match both sides of ReLU with a middle quadratic region. Let $\beta$ to be the \emph{half-width} of a symmetric transition region around $x=0$. Reciprocally to \eref{eq:softplus}-\eref{eq:gelu}, a smaller $\beta$ gives a narrower transition region, and a larger a wider one.  With constraints: a) $y = 0$ to the left of the transition; b) $y = x$ to the right of the transition; c) continuous gradient at the transition boundaries:
\be
 \label{eq:smelu_conditions}
 \left . (dy / dx) \right |_{x=-\beta} = 0,  ~~~ \left . (dy / dx) \right |_{x=\beta} = 1,
\ee
we arrive at the SmeLU activation function:
\be
 \label{eq:smelu}
 y_{\text{SmeLU}} = \left \{
  \begin{array}{rl}
   0; & x \leq -\beta \\
   \frac{(x + \beta)^2}{4\beta}; & |x| \leq \beta \\
    x; & x \geq \beta.
 \end{array}
 \right .
\ee
The 3rd column in Fig.~\ref{fig:smooth} shows SmeLU and its gradient as function of $\beta$.  SmeLU is a convolution of ReLU with a box of magnitude $1/(2\beta)$ in $[-\beta, \beta]$.  Its gradient is a hard Sigmoid.  As we report in Section~\ref{sec:exp}, despite its simplicity, SmeLU's accuracy-reproducibility trade-offs in real systems are competitive with the more complex smooth activations.
A greater $\beta$ tends to produce better PD and a smaller $\beta$ better accuracy.
Using larger $\beta$ values in layers closer to input features and smaller ones closer to the output may allow lowering PDs, because of better smoothness closer to most model parameters (the embeddings) with compensation for accuracy closer to the output.  While this approach interacted well with weight normalization, it did not with layer normalization. 

Figs.~\ref{fig:1d_obj} and~\ref{fig:3d_obj} demonstrate loss landscapes vs. inputs with deep networks whose weights and biases were fixed to random values and a single input is fed into the network (Fig.~\ref{fig:1d_obj}) or two inputs are fed into the network (Fig.~\ref{fig:3d_obj}).  Different loss functions and different normalization strategies are shown.  Activations are clipped within a predefined range.  Curves and surfaces are shown for ReLU, and SmeLU with different $\beta$ values.  The ReLU networks exhibit multiple local (and global) optima, and very non-smooth overall curves or surfaces.  Even with very small $\beta$, smoothing of the curves or surfaces from SmeLU is clear, as well as reduction of the quantities of optima. The objectives appear smoother as we increase $\beta$.  The second row in Fig.~\ref{fig:3d_obj} illustrates the changing of the manifold surrounding a local maximum, which gradually disappears with increasing $\beta$ giving a single global minimum when $\beta = 2$.  Smoothing reduces the opportunities of the optimization to end up in a different minimum.  However, too much smoothing also reduces the ability of the model to distinguish between parameters.  For example, while the loss for $p_1=0.1$ at the top of Fig.~\ref{fig:1d_obj} is clearly different for different inputs for ReLU and SmeLU with $\beta=0.1$, with larger $\beta$ this distinction disappears.  Thus models with large $\beta$ may not be able to distinguish inputs that produce high and low engagement rates.  While it can be more reproducible, its reproducible predictions will have unacceptable accuracy. Using smaller $\beta$, such as $0.1$, on the other hand, may have, in addition to better reproducibility, better accuracy than ReLU because the model is less likely to fall into false local minima.

The surfaces of Fig.~\ref{fig:3d_obj} also demonstrate that the reproducibility improvements of smooth activations do not carry over if initializations are very different between models, or if training example shuffling is too aggressive.  While we see reduction in the quantity of different minima, the surface is still highly non-convex when model accuracy is kept on-par.  If the model starts in different locations, it may not be able to converge to the same minimum.  Aggressive shuffling can affect the trajectory leading to a different region with a different minimum as well.  Thus, smooth activations improve PDs under some locality constraints on the objective surface, as demonstrated in \cite{snapp2021}.  Normalization helps to keep activations within the clipping range.  Otherwise, clipping can destroy the smoothness imposed by the activations.

\subsection{Generalized SmeLU}
The constraints in \eref{eq:smelu_conditions} can be generalized by allowing for asymmetry, shifts, and different gradients
$g_- \neq 0$ to the left and $g_+ > g_-$ to the right of
the quadratic transition region.  Using
\bea
 \nonumber
 \left. (dy / dx) \right |_{x=-\alpha} &=& g_-, \\
 \nonumber
 \left. (dy / dx) \right |_{x=\beta} &=& g_+, \\
 y(x = -\alpha) &=& t
 \label{eq:gsmelu_conditions} 
\eea
\emph{generalized SmeLU (gSmeLU)\/} is defined by $\{\alpha, \beta, g_-, g_+, t\}$.  To cross the origin, $\alpha, \beta > 0$, and $t \leq 0$. For a leaky and monotonic activation, $g_- > 0$.  Enforcing \eref{eq:gsmelu_conditions} and continuity at $x=-\alpha$ and $x=\beta$ leads to
\be
 \label{eq:gsmelu}
 y_{\text{gSmeLU}} = \left \{ \begin{array}{ll}
 g_- x + t + g_- \alpha;& x \leq -\alpha \\
 a x^2 + b x + c; & -\alpha \leq x \leq \beta \\
 g_+ x + t + \frac{\alpha+\beta}{2} g_- + \frac{\alpha -\beta}{2} g_+; & x \geq \beta 
\end{array} \right .
\ee
with
\be
 a = \frac{g_+ - g_-}{2(\alpha + \beta)}, ~
 b = \frac{\alpha g_+ + \beta g_-}{\alpha + \beta} , ~
 c = t + \frac{\alpha^2 ( g_+ + g_-) + 2\alpha\beta g_-}{2(\alpha + \beta)}.
 \label{eq:gsmelu_sup} 
\ee
Examples of gSmeLU (with $t=0$) are shown on the right column of Fig.~\ref{fig:smooth}.
SmeLU is obtained with $g_- = 0$, $g_+=1$, $\alpha = \beta$ and $t=0$.  Hyper-parameters of generalized SmeLU can be learned in training of a deep network together with the model parameters, per network, layer, or unit.  In real systems that optimize for accuracy, layers closer to inputs often learn negative $g_-$, suggesting some level of sparsity regularization.  Closer to the output, $g_-$ closer to $0$ is learned. 
The method of constructing gSmeLU can be extended to use any functions and multiple segments.  We refer to such an activation as \emph{REctified Smooth Continuous Unit (RESCU)}.  (Multiple-piece non-smooth ReLU extensions were considered in \cite{chen18, montufar14}).  

\section{Large-Scale Production Systems}
\label{sec:exp}
SmeLU was deployed in multiple recommendation production systems for multiple products, with a wide range of architectures and objectives, where irreproducibility was a critical concern. For sponsored advertising CTR prediction, using SmeLU allowed replacing costly ensembles by a single network in several products.  In one system, PDs of $17\%$ were observed with single-network ReLU models.  A three component ensemble lowered PD to $12\%$.  SmeLU with a single component network lowered production PDs to $7.5\%$ ($55\%$ reduction) with no accuracy degradation relative to the ensemble, reducing ensemble complexity and technical debt.  Tuning to narrower $\beta$, keeping PD equal to that of the ensemble, gave $0.3\%$ PQAUC loss improvements, giving substantial system metrics improvements.  In another system, SmeLU gave over $0.5\%$ PQAUC loss improvements in production, keeping PDs equal to those of an ensemble, but also providing $33\%$ training speed improvements due to elimination of multiple embedding lookups.  In a different system, PDs were reduced from $20\%$ to $10\%$.  In a content recommendation system, SmeLU provided $6\%$ decrease of recommendation swapping rate, with other additional positive metrics.  In an application installation recommendation system, SmeLU gave up to $6\%$ decrease of PDs. 

\begin{figure}[tbp]
\centerline{
 \includegraphics[width=0.4\textwidth, clip=]{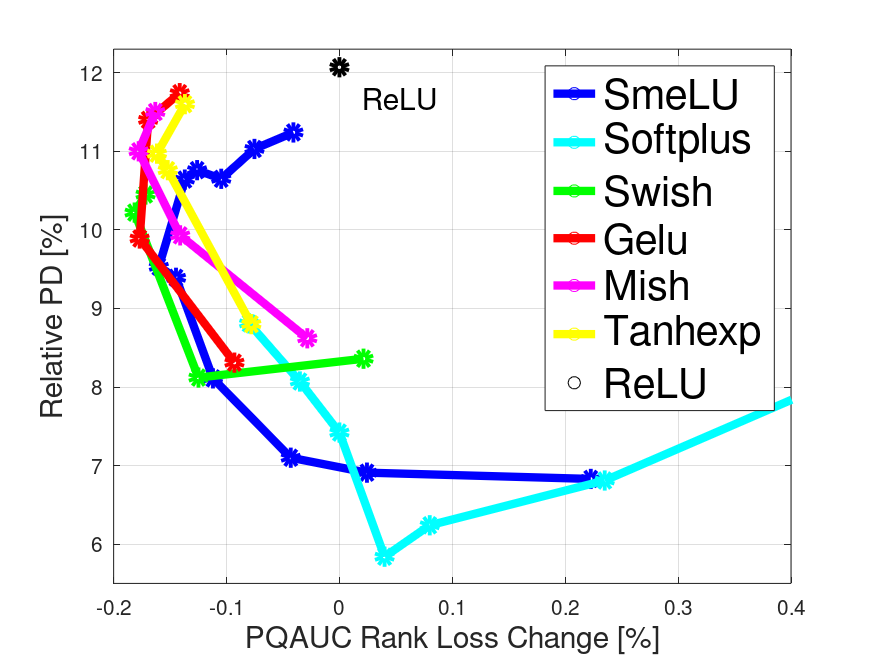}
 }
\caption{Relative PD $\D_r$ in [\%] as a function of PQAUC loss change [\%] from ReLU for different activations on real data (with different $\beta$ values).} 
\label{fig:real_data}
\end{figure}

For a clean comparison of accuracy-reproducibility trade-offs of smooth activations and SmeLU, we also tested them on a ``toy'' simple model on real CTR prediction data for sponsored advertisement.  We constructed a simple $6$ layer fully connected network with dimensions $[1024,512,256,128,64,16]$ with $5$
informative features that were learned as embedding vectors, providing a total of $240$ inputs.  Models were identically initialized, trained with data-shuffling on billions of examples with the system described in Section~\ref{sec:sys}.  Fig.~\ref{fig:real_data} reports relative PD $\D_r$ against  PQAUC ranking loss relative movements.  Losses are expressed as percentage change from a baseline ReLU model.
Moving left implies better accuracy, and down implies
better reproducibility.  For smooth activations, the different points are the result of different $\beta$.  Moving from the left down to the right described increasing $\beta$ for SmeLU (and decreasing for other activations). ReLU has $12\%$ PD.  All smooth activations have points to the left and below ReLU, showing they can be tuned to have both better accuracy and PD.  Differences among smooth activations are subtle. Swish can reach better accuracy, but with worse PD, and Softplus can have better PD, but with bad accuracy.  SmeLU allows lower PDs ($7\%$ on this example) with improved accuracy over ReLU.

Generalized SmeLU gives an additional degrees of freedom over SmeLU. Training the activation together with the model (per layer or unit) led to prediction accuracy improvements.  However, when trained for accuracy, gSmeLU does not give good PDs.  Accuracy improvements can still be leveraged on a teacher model in a distillation set up.  In one system, gSmeLU gave about $0.15\%$ PQAUC loss reduction to the teacher model, which was leveraged to speed up convergence of the student, reducing training time by $13\%$.

\section{Conclusions}
We unveiled the importance of irreproducibility in real world large scale recommendation systems.  We demonstrated how ReLU exacerbates this problem, and described using smooth activations as an approach to mitigate it. We presented SmeLU, which is a simple smooth activation that can be used to improve reproducibility, and reported production launches in real recommendation systems that were able to benefit from SmeLU to improve reproducibility, accuracy, and efficiency.  We also described a generalization of SmeLU that can improve model accuracy, and reported how it was productionalized, gaining efficiency without harming reproducibility.

\section*{Acknowledgement}
The authors acknowledge Sergey Ioffe for early discussions, and Lorenzo Coviello for help in experimentation and launches.

\bibliography{smelu-kdd2022}


\begin{thebibliography}{56}


\ifx \showCODEN    \undefined \def \showCODEN     #1{\unskip}     \fi
\ifx \showDOI      \undefined \def \showDOI       #1{#1}\fi
\ifx \showISBNx    \undefined \def \showISBNx     #1{\unskip}     \fi
\ifx \showISBNxiii \undefined \def \showISBNxiii  #1{\unskip}     \fi
\ifx \showISSN     \undefined \def \showISSN      #1{\unskip}     \fi
\ifx \showLCCN     \undefined \def \showLCCN      #1{\unskip}     \fi
\ifx \shownote     \undefined \def \shownote      #1{#1}          \fi
\ifx \showarticletitle \undefined \def \showarticletitle #1{#1}   \fi
\ifx \showURL      \undefined \def \showURL       {\relax}        \fi
\providecommand\bibfield[2]{#2}
\providecommand\bibinfo[2]{#2}
\providecommand\natexlab[1]{#1}
\providecommand\showeprint[2][]{arXiv:#2}

\bibitem[\protect\citeauthoryear{Achille, Rovere, and Soatto}{Achille
  et~al\mbox{.}}{2017}]%
        {achille17}
\bibfield{author}{\bibinfo{person}{Alessandro Achille}, \bibinfo{person}{Matteo
  Rovere}, {and} \bibinfo{person}{Stefano Soatto}.}
  \bibinfo{year}{2017}\natexlab{}.
\newblock \showarticletitle{Critical learning periods in deep neural networks}.
\newblock \bibinfo{journal}{\emph{arXiv preprint arXiv:1711.08856}}
  (\bibinfo{year}{2017}).
\newblock


\bibitem[\protect\citeauthoryear{Ahn, Jain, Ji, Kale, Netrapalli, and
  Shamir}{Ahn et~al\mbox{.}}{2022}]%
        {ahn21reproducibility}
\bibfield{author}{\bibinfo{person}{Kwangjun Ahn}, \bibinfo{person}{Prateek
  Jain}, \bibinfo{person}{Ziwei Ji}, \bibinfo{person}{Satyen Kale},
  \bibinfo{person}{Praneeth Netrapalli}, {and} \bibinfo{person}{Gil~I.
  Shamir}.} \bibinfo{year}{2022}\natexlab{}.
\newblock \showarticletitle{Reproducibility in optimization: Theoretical
  framework and Limits}.
\newblock \bibinfo{journal}{\emph{arXiv preprint arXiv:2202.04598}}
  (\bibinfo{year}{2022}).
\newblock


\bibitem[\protect\citeauthoryear{Allen-Zhu and Li}{Allen-Zhu and Li}{2020}]%
        {allen2020towards}
\bibfield{author}{\bibinfo{person}{Zeyuan Allen-Zhu} {and}
  \bibinfo{person}{Yuanzhi Li}.} \bibinfo{year}{2020}\natexlab{}.
\newblock \showarticletitle{Towards understanding ensemble, knowledge
  distillation and self-distillation in deep learning}.
\newblock \bibinfo{journal}{\emph{arXiv preprint arXiv:2012.09816}}
  (\bibinfo{year}{2020}).
\newblock


\bibitem[\protect\citeauthoryear{Anil, Pereyra, Passos, Ormandi, Dahl, and
  Hinton}{Anil et~al\mbox{.}}{2018}]%
        {anil18}
\bibfield{author}{\bibinfo{person}{Rohan Anil}, \bibinfo{person}{Gabriel
  Pereyra}, \bibinfo{person}{Alexandre Passos}, \bibinfo{person}{Robert
  Ormandi}, \bibinfo{person}{George~E Dahl}, {and} \bibinfo{person}{Geoffrey~E
  Hinton}.} \bibinfo{year}{2018}\natexlab{}.
\newblock \showarticletitle{Large scale distributed neural network training
  through online distillation}.
\newblock \bibinfo{journal}{\emph{arXiv preprint arXiv:1804.03235}}
  (\bibinfo{year}{2018}).
\newblock


\bibitem[\protect\citeauthoryear{Ba, Kiros, and Hinton}{Ba
  et~al\mbox{.}}{2016}]%
        {ba16}
\bibfield{author}{\bibinfo{person}{Jimmy~Lei Ba}, \bibinfo{person}{Jamie~Ryan
  Kiros}, {and} \bibinfo{person}{Geoffrey~E Hinton}.}
  \bibinfo{year}{2016}\natexlab{}.
\newblock \showarticletitle{Layer normalization}.
\newblock \bibinfo{journal}{\emph{arXiv preprint arXiv:1607.06450}}
  (\bibinfo{year}{2016}).
\newblock


\bibitem[\protect\citeauthoryear{Barron}{Barron}{2017}]%
        {barron17}
\bibfield{author}{\bibinfo{person}{Jonathan~T Barron}.}
  \bibinfo{year}{2017}\natexlab{}.
\newblock \showarticletitle{Continuously differentiable exponential linear
  units}.
\newblock \bibinfo{journal}{\emph{arXiv}} (\bibinfo{year}{2017}),
  \bibinfo{pages}{arXiv--1704}.
\newblock


\bibitem[\protect\citeauthoryear{Bhojanapalli, Wilber, Veit, Rawat, Kim, Menon,
  and Kumar}{Bhojanapalli et~al\mbox{.}}{2021}]%
        {bho21}
\bibfield{author}{\bibinfo{person}{Srinadh Bhojanapalli},
  \bibinfo{person}{Kimberly~Jenney Wilber}, \bibinfo{person}{Andreas Veit},
  \bibinfo{person}{Ankit~Singh Rawat}, \bibinfo{person}{Seungyeon Kim},
  \bibinfo{person}{Aditya~Krishna Menon}, {and} \bibinfo{person}{Sanjiv
  Kumar}.} \bibinfo{year}{2021}\natexlab{}.
\newblock \bibinfo{title}{On the Reproducibility of Neural Network
  Predictions}.
\newblock
\newblock


\bibitem[\protect\citeauthoryear{Blum, Kalai, and Langford}{Blum
  et~al\mbox{.}}{1999}]%
        {blum99}
\bibfield{author}{\bibinfo{person}{Avrim Blum}, \bibinfo{person}{Adam Kalai},
  {and} \bibinfo{person}{John Langford}.} \bibinfo{year}{1999}\natexlab{}.
\newblock \showarticletitle{Beating the hold-out: Bounds for k-fold and
  progressive cross-validation}. In \bibinfo{booktitle}{\emph{Proceedings of
  the twelfth annual conference on Computational learning theory}}.
  \bibinfo{pages}{203--208}.
\newblock


\bibitem[\protect\citeauthoryear{Bresler and Nagaraj}{Bresler and
  Nagaraj}{2020}]%
        {bresler20}
\bibfield{author}{\bibinfo{person}{Guy Bresler} {and} \bibinfo{person}{Dheeraj
  Nagaraj}.} \bibinfo{year}{2020}\natexlab{}.
\newblock \showarticletitle{A Corrective View of Neural Networks:
  Representation, Memorization and Learning}. In \bibinfo{booktitle}{\emph{COLT
  2020}} \emph{(\bibinfo{series}{Proceedings of Machine Learning Research},
  Vol.~\bibinfo{volume}{125})}, \bibfield{editor}{\bibinfo{person}{Jacob
  Abernethy} {and} \bibinfo{person}{Shivani Agarwal}} (Eds.).
  \bibinfo{publisher}{PMLR}, \bibinfo{pages}{848--901}.
\newblock


\bibitem[\protect\citeauthoryear{Chen and Ho}{Chen and Ho}{2018}]%
        {chen18}
\bibfield{author}{\bibinfo{person}{Zhi Chen} {and} \bibinfo{person}{Pin-han
  Ho}.} \bibinfo{year}{2018}\natexlab{}.
\newblock \showarticletitle{Deep Global-Connected Net With The Generalized
  Multi-Piecewise ReLU Activation in Deep Learning}.
\newblock \bibinfo{journal}{\emph{arXiv preprint arXiv:1807.03116}}
  (\bibinfo{year}{2018}).
\newblock


\bibitem[\protect\citeauthoryear{Chen, Wang, Lin, Cheng, Hong, Chi, and
  Cui}{Chen et~al\mbox{.}}{2020}]%
        {chen20}
\bibfield{author}{\bibinfo{person}{Zhe Chen}, \bibinfo{person}{Yuyan Wang},
  \bibinfo{person}{Dong Lin}, \bibinfo{person}{Derek Cheng},
  \bibinfo{person}{Lichan Hong}, \bibinfo{person}{Ed Chi}, {and}
  \bibinfo{person}{Claire Cui}.} \bibinfo{year}{2020}\natexlab{}.
\newblock \showarticletitle{Beyond point estimate: Inferring ensemble
  prediction variation from neuron activation strength in recommender systems}.
\newblock \bibinfo{journal}{\emph{arXiv preprint arXiv:2008.07032}}
  (\bibinfo{year}{2020}).
\newblock


\bibitem[\protect\citeauthoryear{Clevert, Unterthiner, and Hochreiter}{Clevert
  et~al\mbox{.}}{2015}]%
        {clevert15}
\bibfield{author}{\bibinfo{person}{Djork-Arn{\'e} Clevert},
  \bibinfo{person}{Thomas Unterthiner}, {and} \bibinfo{person}{Sepp
  Hochreiter}.} \bibinfo{year}{2015}\natexlab{}.
\newblock \showarticletitle{Fast and accurate deep network learning by
  exponential linear units (elus)}.
\newblock \bibinfo{journal}{\emph{arXiv preprint arXiv:1511.07289}}
  (\bibinfo{year}{2015}).
\newblock


\bibitem[\protect\citeauthoryear{D'Amour, Heller, Moldovan, Adlam, Alipanahi,
  Beutel, Chen, Deaton, Eisenstein, Hoffman, Hormozdiari, Houlsby, Hou, Jerfel,
  Karthikesalingam, Lucic, Ma, McLean, Mincu, Mitani, Montanari, Nado,
  Natarajan, Nielson, Osborne, Raman, Ramasamy, Sayres, Schrouff, Seneviratne,
  Sequeira, Suresh, Veitch, Vladymyrov, Wang, Webster, Yadlowsky, Yun, Zhai,
  and Sculley}{D'Amour et~al\mbox{.}}{2020}]%
        {damour20}
\bibfield{author}{\bibinfo{person}{Alexander D'Amour},
  \bibinfo{person}{Katherine Heller}, \bibinfo{person}{Dan Moldovan},
  \bibinfo{person}{Ben Adlam}, \bibinfo{person}{Babak Alipanahi},
  \bibinfo{person}{Alex Beutel}, \bibinfo{person}{Christina Chen},
  \bibinfo{person}{Jonathan Deaton}, \bibinfo{person}{Jacob Eisenstein},
  \bibinfo{person}{Matthew~D. Hoffman}, \bibinfo{person}{Farhad Hormozdiari},
  \bibinfo{person}{Neil Houlsby}, \bibinfo{person}{Shaobo Hou},
  \bibinfo{person}{Ghassen Jerfel}, \bibinfo{person}{Alan Karthikesalingam},
  \bibinfo{person}{Mario Lucic}, \bibinfo{person}{Yian Ma},
  \bibinfo{person}{Cory McLean}, \bibinfo{person}{Diana Mincu},
  \bibinfo{person}{Akinori Mitani}, \bibinfo{person}{Andrea Montanari},
  \bibinfo{person}{Zachary Nado}, \bibinfo{person}{Vivek Natarajan},
  \bibinfo{person}{Christopher Nielson}, \bibinfo{person}{Thomas~F. Osborne},
  \bibinfo{person}{Rajiv Raman}, \bibinfo{person}{Kim Ramasamy},
  \bibinfo{person}{Rory Sayres}, \bibinfo{person}{Jessica Schrouff},
  \bibinfo{person}{Martin Seneviratne}, \bibinfo{person}{Shannon Sequeira},
  \bibinfo{person}{Harini Suresh}, \bibinfo{person}{Victor Veitch},
  \bibinfo{person}{Max Vladymyrov}, \bibinfo{person}{Xuezhi Wang},
  \bibinfo{person}{Kellie Webster}, \bibinfo{person}{Steve Yadlowsky},
  \bibinfo{person}{Taedong Yun}, \bibinfo{person}{Xiaohua Zhai}, {and}
  \bibinfo{person}{D. Sculley}.} \bibinfo{year}{2020}\natexlab{}.
\newblock \bibinfo{title}{Underspecification Presents Challenges for
  Credibility in Modern Machine Learning}.
\newblock
\newblock
\showeprint[arxiv]{2011.03395}~[cs.LG]


\bibitem[\protect\citeauthoryear{Dietterich}{Dietterich}{2000}]%
        {dietterich00}
\bibfield{author}{\bibinfo{person}{T.~G. Dietterich}.}
  \bibinfo{year}{2000}\natexlab{}.
\newblock \showarticletitle{Ensemble methods in machine learning}.
\newblock \bibinfo{journal}{\emph{Lecture Notes in Computer Science}}
  (\bibinfo{year}{2000}), \bibinfo{pages}{1--15}.
\newblock


\bibitem[\protect\citeauthoryear{Du}{Du}{2019}]%
        {du19}
\bibfield{author}{\bibinfo{person}{Simon Du}.} \bibinfo{year}{2019}\natexlab{}.
\newblock \emph{\bibinfo{title}{Gradient Descent for Non-convex Problems in
  Modern Machine Learning}}.
\newblock \bibinfo{thesistype}{Ph.\,D. Dissertation}. \bibinfo{school}{Carnegie
  Mellon University}.
\newblock


\bibitem[\protect\citeauthoryear{Duchi, Hazan, and Singer}{Duchi
  et~al\mbox{.}}{2011}]%
        {duchi11}
\bibfield{author}{\bibinfo{person}{J. Duchi}, \bibinfo{person}{E. Hazan}, {and}
  \bibinfo{person}{Y. Singer}.} \bibinfo{year}{2011}\natexlab{}.
\newblock \showarticletitle{Adaptive subgradient methods for online learning
  and stochastic optimization}.
\newblock \bibinfo{journal}{\emph{Journal of Machine Learning Research}}
  \bibinfo{volume}{12} (\bibinfo{date}{Feb.} \bibinfo{year}{2011}),
  \bibinfo{pages}{2121--2159}.
\newblock


\bibitem[\protect\citeauthoryear{Dusenberry, Tran, Choi, Kemp, Nixon, Jerfel,
  Heller, and Dai}{Dusenberry et~al\mbox{.}}{2020}]%
        {dusenberry20}
\bibfield{author}{\bibinfo{person}{Michael~W Dusenberry},
  \bibinfo{person}{Dustin Tran}, \bibinfo{person}{Edward Choi},
  \bibinfo{person}{Jonas Kemp}, \bibinfo{person}{Jeremy Nixon},
  \bibinfo{person}{Ghassen Jerfel}, \bibinfo{person}{Katherine Heller}, {and}
  \bibinfo{person}{Andrew~M Dai}.} \bibinfo{year}{2020}\natexlab{}.
\newblock \showarticletitle{Analyzing the role of model uncertainty for
  electronic health records}. In \bibinfo{booktitle}{\emph{Proceedings of the
  ACM Conference on Health, Inference, and Learning}}.
  \bibinfo{pages}{204--213}.
\newblock


\bibitem[\protect\citeauthoryear{Fort, Hu, and Lakshminarayanan}{Fort
  et~al\mbox{.}}{2020}]%
        {fort2020deep}
\bibfield{author}{\bibinfo{person}{Stanislav Fort}, \bibinfo{person}{Huiyi Hu},
  {and} \bibinfo{person}{Balaji Lakshminarayanan}.}
  \bibinfo{year}{2020}\natexlab{}.
\newblock \bibinfo{title}{Deep Ensembles: A Loss Landscape Perspective}.
\newblock
\newblock
\showeprint[arxiv]{1912.02757}~[stat.ML]


\bibitem[\protect\citeauthoryear{Frankle, Dziugaite, Roy, and Carbin}{Frankle
  et~al\mbox{.}}{2020}]%
        {frankle2020linear}
\bibfield{author}{\bibinfo{person}{Jonathan Frankle},
  \bibinfo{person}{Gintare~Karolina Dziugaite}, \bibinfo{person}{Daniel Roy},
  {and} \bibinfo{person}{Michael Carbin}.} \bibinfo{year}{2020}\natexlab{}.
\newblock \showarticletitle{Linear mode connectivity and the lottery ticket
  hypothesis}. In \bibinfo{booktitle}{\emph{International Conference on Machine
  Learning}}. PMLR, \bibinfo{pages}{3259--3269}.
\newblock


\bibitem[\protect\citeauthoryear{Hendrycks and Gimpel}{Hendrycks and
  Gimpel}{2016}]%
        {hendrycks16}
\bibfield{author}{\bibinfo{person}{Dan Hendrycks} {and} \bibinfo{person}{Kevin
  Gimpel}.} \bibinfo{year}{2016}\natexlab{}.
\newblock \showarticletitle{Gaussian error linear units (gelus)}.
\newblock \bibinfo{journal}{\emph{arXiv preprint arXiv:1606.08415}}
  (\bibinfo{year}{2016}).
\newblock


\bibitem[\protect\citeauthoryear{Hinton, Vinyals, and Dean}{Hinton
  et~al\mbox{.}}{2015}]%
        {hinton15}
\bibfield{author}{\bibinfo{person}{Geoffrey Hinton}, \bibinfo{person}{Oriol
  Vinyals}, {and} \bibinfo{person}{Jeff Dean}.}
  \bibinfo{year}{2015}\natexlab{}.
\newblock \showarticletitle{Distilling the knowledge in a neural network}.
\newblock \bibinfo{journal}{\emph{arXiv preprint arXiv:1503.02531}}
  (\bibinfo{year}{2015}).
\newblock


\bibitem[\protect\citeauthoryear{Huber}{Huber}{1992}]%
        {huber92}
\bibfield{author}{\bibinfo{person}{Peter~J Huber}.}
  \bibinfo{year}{1992}\natexlab{}.
\newblock \showarticletitle{Robust estimation of a location parameter}.
\newblock In \bibinfo{booktitle}{\emph{Breakthroughs in statistics}}.
  \bibinfo{publisher}{Springer}, \bibinfo{pages}{492--518}.
\newblock


\bibitem[\protect\citeauthoryear{Ioffe and Szegedy}{Ioffe and Szegedy}{2015}]%
        {ioffe15}
\bibfield{author}{\bibinfo{person}{Sergey Ioffe} {and}
  \bibinfo{person}{Christian Szegedy}.} \bibinfo{year}{2015}\natexlab{}.
\newblock \showarticletitle{Batch normalization: Accelerating deep network
  training by reducing internal covariate shift}.
\newblock \bibinfo{journal}{\emph{arXiv preprint arXiv:1502.03167}}
  (\bibinfo{year}{2015}).
\newblock


\bibitem[\protect\citeauthoryear{Jin, Xu, Feng, Wei, Xiong, and Yan}{Jin
  et~al\mbox{.}}{2015}]%
        {jin15}
\bibfield{author}{\bibinfo{person}{Xiaojie Jin}, \bibinfo{person}{Chunyan Xu},
  \bibinfo{person}{Jiashi Feng}, \bibinfo{person}{Yunchao Wei},
  \bibinfo{person}{Junjun Xiong}, {and} \bibinfo{person}{Shuicheng Yan}.}
  \bibinfo{year}{2015}\natexlab{}.
\newblock \showarticletitle{Deep learning with s-shaped rectified linear
  activation units}.
\newblock \bibinfo{journal}{\emph{arXiv preprint arXiv:1512.07030}}
  (\bibinfo{year}{2015}).
\newblock


\bibitem[\protect\citeauthoryear{Klambauer, Unterthiner, Mayr, and
  Hochreiter}{Klambauer et~al\mbox{.}}{2017}]%
        {klambauer17}
\bibfield{author}{\bibinfo{person}{G{\"u}nter Klambauer},
  \bibinfo{person}{Thomas Unterthiner}, \bibinfo{person}{Andreas Mayr}, {and}
  \bibinfo{person}{Sepp Hochreiter}.} \bibinfo{year}{2017}\natexlab{}.
\newblock \showarticletitle{Self-normalizing neural networks}. In
  \bibinfo{booktitle}{\emph{Advances in neural information processing
  systems}}. \bibinfo{pages}{971--980}.
\newblock


\bibitem[\protect\citeauthoryear{Kondratyuk, Tan, Brown, and Gong}{Kondratyuk
  et~al\mbox{.}}{2020}]%
        {kondratyuk2020ensembling}
\bibfield{author}{\bibinfo{person}{Dan Kondratyuk}, \bibinfo{person}{Mingxing
  Tan}, \bibinfo{person}{Matthew Brown}, {and} \bibinfo{person}{Boqing Gong}.}
  \bibinfo{year}{2020}\natexlab{}.
\newblock \showarticletitle{When ensembling smaller models is more efficient
  than single large models}.
\newblock \bibinfo{journal}{\emph{arXiv preprint arXiv:2005.00570}}
  (\bibinfo{year}{2020}).
\newblock


\bibitem[\protect\citeauthoryear{Lakshminarayanan, Pritzel, and
  Blundell}{Lakshminarayanan et~al\mbox{.}}{2017}]%
        {lakshminarayanan17}
\bibfield{author}{\bibinfo{person}{Balaji Lakshminarayanan},
  \bibinfo{person}{Alexander Pritzel}, {and} \bibinfo{person}{Charles
  Blundell}.} \bibinfo{year}{2017}\natexlab{}.
\newblock \showarticletitle{Simple and scalable predictive uncertainty
  estimation using deep ensembles}. In \bibinfo{booktitle}{\emph{Advances in
  neural information processing systems}}. \bibinfo{pages}{6402--6413}.
\newblock


\bibitem[\protect\citeauthoryear{Liu and Di}{Liu and Di}{2020}]%
        {liu20tanhexp}
\bibfield{author}{\bibinfo{person}{Xinyu Liu} {and} \bibinfo{person}{Xiaoguang
  Di}.} \bibinfo{year}{2020}\natexlab{}.
\newblock \showarticletitle{TanhExp: A Smooth Activation Function with High
  Convergence Speed for Lightweight Neural Networks}.
\newblock \bibinfo{journal}{\emph{arXiv preprint arXiv:2003.09855}}
  (\bibinfo{year}{2020}).
\newblock


\bibitem[\protect\citeauthoryear{Lobacheva, Chirkova, Kodryan, and
  Vetrov}{Lobacheva et~al\mbox{.}}{2020}]%
        {lobacheva2020power}
\bibfield{author}{\bibinfo{person}{Ekaterina Lobacheva},
  \bibinfo{person}{Nadezhda Chirkova}, \bibinfo{person}{Maxim Kodryan}, {and}
  \bibinfo{person}{Dmitry Vetrov}.} \bibinfo{year}{2020}\natexlab{}.
\newblock \showarticletitle{On power laws in deep ensembles}.
\newblock \bibinfo{journal}{\emph{arXiv preprint arXiv:2007.08483}}
  (\bibinfo{year}{2020}).
\newblock


\bibitem[\protect\citeauthoryear{Lokhande, Tasneeyapant, Venkatesh, Ravi, and
  Singh}{Lokhande et~al\mbox{.}}{2020}]%
        {lokhande20}
\bibfield{author}{\bibinfo{person}{Vishnu~Suresh Lokhande},
  \bibinfo{person}{Songwong Tasneeyapant}, \bibinfo{person}{Abhay Venkatesh},
  \bibinfo{person}{Sathya~N Ravi}, {and} \bibinfo{person}{Vikas Singh}.}
  \bibinfo{year}{2020}\natexlab{}.
\newblock \showarticletitle{Generating Accurate Pseudo-Labels in
  Semi-Supervised Learning and Avoiding Overconfident Predictions via Hermite
  Polynomial Activations}. In \bibinfo{booktitle}{\emph{Proceedings of the
  IEEE/CVF Conference on Computer Vision and Pattern Recognition}}.
  \bibinfo{pages}{11435--11443}.
\newblock


\bibitem[\protect\citeauthoryear{McMahan, Holt, Sculley, Young, Ebner, Grady,
  Nie, Phillips, Davydov, Golovin, et~al\mbox{.}}{McMahan
  et~al\mbox{.}}{2013}]%
        {mcmahan13}
\bibfield{author}{\bibinfo{person}{H~Brendan McMahan}, \bibinfo{person}{Gary
  Holt}, \bibinfo{person}{David Sculley}, \bibinfo{person}{Michael Young},
  \bibinfo{person}{Dietmar Ebner}, \bibinfo{person}{Julian Grady},
  \bibinfo{person}{Lan Nie}, \bibinfo{person}{Todd Phillips},
  \bibinfo{person}{Eugene Davydov}, \bibinfo{person}{Daniel Golovin},
  {et~al\mbox{.}}} \bibinfo{year}{2013}\natexlab{}.
\newblock \showarticletitle{Ad click prediction: a view from the trenches}. In
  \bibinfo{booktitle}{\emph{Proceedings of the 19th ACM SIGKDD international
  conference on Knowledge discovery and data mining}}.
  \bibinfo{pages}{1222--1230}.
\newblock


\bibitem[\protect\citeauthoryear{Mhaskar}{Mhaskar}{1997}]%
        {mhaskar97}
\bibfield{author}{\bibinfo{person}{HN Mhaskar}.}
  \bibinfo{year}{1997}\natexlab{}.
\newblock \showarticletitle{On smooth activation functions}.
\newblock In \bibinfo{booktitle}{\emph{Mathematics of Neural Networks}}.
  \bibinfo{publisher}{Springer}, \bibinfo{pages}{275--279}.
\newblock


\bibitem[\protect\citeauthoryear{Misra}{Misra}{2019}]%
        {misra19}
\bibfield{author}{\bibinfo{person}{Diganta Misra}.}
  \bibinfo{year}{2019}\natexlab{}.
\newblock \showarticletitle{Mish: A self regularized non-monotonic neural
  activation function}.
\newblock \bibinfo{journal}{\emph{arXiv preprint arXiv:1908.08681}}
  (\bibinfo{year}{2019}).
\newblock


\bibitem[\protect\citeauthoryear{Montufar, Pascanu, Cho, and Bengio}{Montufar
  et~al\mbox{.}}{2014}]%
        {montufar14}
\bibfield{author}{\bibinfo{person}{Guido~F Montufar}, \bibinfo{person}{Razvan
  Pascanu}, \bibinfo{person}{Kyunghyun Cho}, {and} \bibinfo{person}{Yoshua
  Bengio}.} \bibinfo{year}{2014}\natexlab{}.
\newblock \showarticletitle{On the number of linear regions of deep neural
  networks}. In \bibinfo{booktitle}{\emph{Advances in neural information
  processing systems}}. \bibinfo{pages}{2924--2932}.
\newblock


\bibitem[\protect\citeauthoryear{Nagarajan, Warnell, and Stone}{Nagarajan
  et~al\mbox{.}}{2018}]%
        {nagarajan18}
\bibfield{author}{\bibinfo{person}{Prabhat Nagarajan}, \bibinfo{person}{Garrett
  Warnell}, {and} \bibinfo{person}{Peter Stone}.}
  \bibinfo{year}{2018}\natexlab{}.
\newblock \showarticletitle{Deterministic implementations for reproducibility
  in deep reinforcement learning}.
\newblock \bibinfo{journal}{\emph{arXiv preprint arXiv:1809.05676}}
  (\bibinfo{year}{2018}).
\newblock


\bibitem[\protect\citeauthoryear{Nair and Hinton}{Nair and Hinton}{2010}]%
        {nair10}
\bibfield{author}{\bibinfo{person}{Vinod Nair} {and}
  \bibinfo{person}{Geoffrey~E Hinton}.} \bibinfo{year}{2010}\natexlab{}.
\newblock \showarticletitle{Rectified linear units improve restricted boltzmann
  machines}. In \bibinfo{booktitle}{\emph{ICML}}.
\newblock


\bibitem[\protect\citeauthoryear{Nwankpa, Ijomah, Gachagan, and
  Marshall}{Nwankpa et~al\mbox{.}}{2018}]%
        {nwankpa18}
\bibfield{author}{\bibinfo{person}{Chigozie Nwankpa}, \bibinfo{person}{Winifred
  Ijomah}, \bibinfo{person}{Anthony Gachagan}, {and} \bibinfo{person}{Stephen
  Marshall}.} \bibinfo{year}{2018}\natexlab{}.
\newblock \showarticletitle{Activation functions: Comparison of trends in
  practice and research for deep learning}.
\newblock \bibinfo{journal}{\emph{arXiv preprint arXiv:1811.03378}}
  (\bibinfo{year}{2018}).
\newblock


\bibitem[\protect\citeauthoryear{Pedamonti}{Pedamonti}{2018}]%
        {pedamonti18}
\bibfield{author}{\bibinfo{person}{Dabal Pedamonti}.}
  \bibinfo{year}{2018}\natexlab{}.
\newblock \showarticletitle{Comparison of non-linear activation functions for
  deep neural networks on MNIST classification task}.
\newblock \bibinfo{journal}{\emph{arXiv preprint arXiv:1804.02763}}
  (\bibinfo{year}{2018}).
\newblock


\bibitem[\protect\citeauthoryear{Pineau, Vincent-Lamarre, Sinha, Larivi{\`e}re,
  Beygelzimer, d’Alch{\'e} Buc, Fox, and Larochelle}{Pineau
  et~al\mbox{.}}{2021}]%
        {pineau2021improving}
\bibfield{author}{\bibinfo{person}{Joelle Pineau}, \bibinfo{person}{Philippe
  Vincent-Lamarre}, \bibinfo{person}{Koustuv Sinha}, \bibinfo{person}{Vincent
  Larivi{\`e}re}, \bibinfo{person}{Alina Beygelzimer},
  \bibinfo{person}{Florence d’Alch{\'e} Buc}, \bibinfo{person}{Emily Fox},
  {and} \bibinfo{person}{Hugo Larochelle}.} \bibinfo{year}{2021}\natexlab{}.
\newblock \showarticletitle{Improving reproducibility in machine learning
  research: a report from the NeurIPS 2019 reproducibility program}.
\newblock \bibinfo{journal}{\emph{Journal of Machine Learning Research}}
  \bibinfo{volume}{22} (\bibinfo{year}{2021}).
\newblock


\bibitem[\protect\citeauthoryear{Ramachandran, Zoph, and Le}{Ramachandran
  et~al\mbox{.}}{2017}]%
        {ramachandran17}
\bibfield{author}{\bibinfo{person}{Prajit Ramachandran},
  \bibinfo{person}{Barret Zoph}, {and} \bibinfo{person}{Quoc~V Le}.}
  \bibinfo{year}{2017}\natexlab{}.
\newblock \showarticletitle{Searching for activation functions}.
\newblock \bibinfo{journal}{\emph{arXiv preprint arXiv:1710.05941}}
  (\bibinfo{year}{2017}).
\newblock


\bibitem[\protect\citeauthoryear{Sakketou and Ampazis}{Sakketou and
  Ampazis}{2019}]%
        {sakketou19}
\bibfield{author}{\bibinfo{person}{Flora Sakketou} {and}
  \bibinfo{person}{Nicholas Ampazis}.} \bibinfo{year}{2019}\natexlab{}.
\newblock \showarticletitle{On the Invariance of the SELU Activation Function
  on Algorithm and Hyperparameter Selection in Neural Network Recommenders}. In
  \bibinfo{booktitle}{\emph{IFIP International Conference on Artificial
  Intelligence Applications and Innovations}}. Springer,
  \bibinfo{pages}{673--685}.
\newblock


\bibitem[\protect\citeauthoryear{Salimans and Kingma}{Salimans and
  Kingma}{2016}]%
        {salimans16}
\bibfield{author}{\bibinfo{person}{Tim Salimans} {and} \bibinfo{person}{Durk~P
  Kingma}.} \bibinfo{year}{2016}\natexlab{}.
\newblock \showarticletitle{Weight normalization: A simple reparameterization
  to accelerate training of deep neural networks}. In
  \bibinfo{booktitle}{\emph{Advances in neural information processing
  systems}}. \bibinfo{pages}{901--909}.
\newblock


\bibitem[\protect\citeauthoryear{Sculley, Holt, Golovin, Davydov, Phillips,
  Ebner, Chaudhary, and Young}{Sculley et~al\mbox{.}}{2014}]%
        {sculley14}
\bibfield{author}{\bibinfo{person}{David Sculley}, \bibinfo{person}{Gary Holt},
  \bibinfo{person}{Daniel Golovin}, \bibinfo{person}{Eugene Davydov},
  \bibinfo{person}{Todd Phillips}, \bibinfo{person}{Dietmar Ebner},
  \bibinfo{person}{Vinay Chaudhary}, {and} \bibinfo{person}{Michael Young}.}
  \bibinfo{year}{2014}\natexlab{}.
\newblock \showarticletitle{Machine learning: The high interest credit card of
  technical debt}.
\newblock  (\bibinfo{year}{2014}).
\newblock


\bibitem[\protect\citeauthoryear{Shallue, Lee, Antognini, Sohl-Dickstein,
  Frostig, and Dahl}{Shallue et~al\mbox{.}}{2018}]%
        {shallue18}
\bibfield{author}{\bibinfo{person}{Christopher~J Shallue},
  \bibinfo{person}{Jaehoon Lee}, \bibinfo{person}{Joseph Antognini},
  \bibinfo{person}{Jascha Sohl-Dickstein}, \bibinfo{person}{Roy Frostig}, {and}
  \bibinfo{person}{George~E Dahl}.} \bibinfo{year}{2018}\natexlab{}.
\newblock \showarticletitle{Measuring the effects of data parallelism on neural
  network training}.
\newblock \bibinfo{journal}{\emph{arXiv preprint arXiv:1811.03600}}
  (\bibinfo{year}{2018}).
\newblock


\bibitem[\protect\citeauthoryear{Shamir}{Shamir}{2018}]%
        {shamir18}
\bibfield{author}{\bibinfo{person}{Gil~I. Shamir}.}
  \bibinfo{year}{2018}\natexlab{}.
\newblock \bibinfo{title}{Systems and Methods for Improved Generalization,
  Reproducibility, and Stabilization of Neural Networks via Error Control Code
  Constraints}.
\newblock
\newblock


\bibitem[\protect\citeauthoryear{Shamir and Coviello}{Shamir and
  Coviello}{2020a}]%
        {shamir20}
\bibfield{author}{\bibinfo{person}{Gil~I Shamir} {and} \bibinfo{person}{Lorenzo
  Coviello}.} \bibinfo{year}{2020}\natexlab{a}.
\newblock \showarticletitle{Anti-Distillation: Improving Reproducibility of
  Deep Networks}.
\newblock \bibinfo{journal}{\emph{Preprint}} (\bibinfo{year}{2020}).
\newblock


\bibitem[\protect\citeauthoryear{Shamir and Coviello}{Shamir and
  Coviello}{2020b}]%
        {shamir20ed}
\bibfield{author}{\bibinfo{person}{Gil~I. Shamir} {and}
  \bibinfo{person}{Lorenzo Coviello}.} \bibinfo{year}{2020}\natexlab{b}.
\newblock \bibinfo{title}{Distilling from Ensembles to Improve Reproducibility
  of Neural Networks}.
\newblock
\newblock


\bibitem[\protect\citeauthoryear{Snapp and Shamir}{Snapp and Shamir}{2021}]%
        {snapp2021}
\bibfield{author}{\bibinfo{person}{Robert~R Snapp} {and} \bibinfo{person}{Gil~I
  Shamir}.} \bibinfo{year}{2021}\natexlab{}.
\newblock \showarticletitle{Synthesizing Irreproducibility in Deep Networks}.
\newblock \bibinfo{journal}{\emph{arXiv preprint arXiv:2102.10696}}
  (\bibinfo{year}{2021}).
\newblock


\bibitem[\protect\citeauthoryear{Summers and Dinneen}{Summers and
  Dinneen}{2021}]%
        {summers21}
\bibfield{author}{\bibinfo{person}{Cecilia Summers} {and}
  \bibinfo{person}{Michael~J. Dinneen}.} \bibinfo{year}{2021}\natexlab{}.
\newblock \bibinfo{title}{On Nondeterminism and Instability in Neural Network
  Optimization}.
\newblock
\newblock


\bibitem[\protect\citeauthoryear{Wang, Qin, and Zhu}{Wang
  et~al\mbox{.}}{2017}]%
        {wang17}
\bibfield{author}{\bibinfo{person}{Tianyang Wang}, \bibinfo{person}{Zhengrui
  Qin}, {and} \bibinfo{person}{Michelle Zhu}.} \bibinfo{year}{2017}\natexlab{}.
\newblock \showarticletitle{An ELU network with total variation for image
  denoising}. In \bibinfo{booktitle}{\emph{International Conference on Neural
  Information Processing}}. Springer, \bibinfo{pages}{227--237}.
\newblock


\bibitem[\protect\citeauthoryear{Wang, Kondratyuk, Christiansen, Kitani, Alon,
  and Eban}{Wang et~al\mbox{.}}{2021}]%
        {wang2021wisdom}
\bibfield{author}{\bibinfo{person}{Xiaofang Wang}, \bibinfo{person}{Dan
  Kondratyuk}, \bibinfo{person}{Eric Christiansen}, \bibinfo{person}{Kris~M.
  Kitani}, \bibinfo{person}{Yair Alon}, {and} \bibinfo{person}{Elad Eban}.}
  \bibinfo{year}{2021}\natexlab{}.
\newblock \showarticletitle{Wisdom of Committees: An Overlooked Approach To
  Faster and More Accurate Models}.
\newblock \bibinfo{journal}{\emph{arXiv preprint arXiv:2012.01988}}
  (\bibinfo{year}{2021}).
\newblock


\bibitem[\protect\citeauthoryear{Xie, Tan, Gong, Yuille, and Le}{Xie
  et~al\mbox{.}}{2020}]%
        {xie20}
\bibfield{author}{\bibinfo{person}{Cihang Xie}, \bibinfo{person}{Mingxing Tan},
  \bibinfo{person}{Boqing Gong}, \bibinfo{person}{Alan Yuille}, {and}
  \bibinfo{person}{Quoc~V Le}.} \bibinfo{year}{2020}\natexlab{}.
\newblock \showarticletitle{Smooth adversarial training}.
\newblock \bibinfo{journal}{\emph{arXiv preprint arXiv:2006.14536}}
  (\bibinfo{year}{2020}).
\newblock


\bibitem[\protect\citeauthoryear{Yu, Chen, Lin, Shamir, and Han}{Yu
  et~al\mbox{.}}{2021}]%
        {yu2021dropout}
\bibfield{author}{\bibinfo{person}{Haichao Yu}, \bibinfo{person}{Zhe Chen},
  \bibinfo{person}{Dong Lin}, \bibinfo{person}{Gil Shamir}, {and}
  \bibinfo{person}{Jie Han}.} \bibinfo{year}{2021}\natexlab{}.
\newblock \showarticletitle{Dropout Prediction Variation Estimation Using
  Neuron Activation Strength}.
\newblock \bibinfo{journal}{\emph{arXiv preprint arXiv:2110.06435}}
  (\bibinfo{year}{2021}).
\newblock


\bibitem[\protect\citeauthoryear{Zhang, Xiang, Hospedales, and Lu}{Zhang
  et~al\mbox{.}}{2018}]%
        {zhang18}
\bibfield{author}{\bibinfo{person}{Ying Zhang}, \bibinfo{person}{Tao Xiang},
  \bibinfo{person}{Timothy~M Hospedales}, {and} \bibinfo{person}{Huchuan Lu}.}
  \bibinfo{year}{2018}\natexlab{}.
\newblock \showarticletitle{Deep mutual learning}. In
  \bibinfo{booktitle}{\emph{Proceedings of the IEEE Conference on Computer
  Vision and Pattern Recognition}}. \bibinfo{pages}{4320--4328}.
\newblock


\bibitem[\protect\citeauthoryear{Zheng, Yang, Liu, Liang, and Li}{Zheng
  et~al\mbox{.}}{2015}]%
        {zheng15}
\bibfield{author}{\bibinfo{person}{Hao Zheng}, \bibinfo{person}{Zhanlei Yang},
  \bibinfo{person}{Wenju Liu}, \bibinfo{person}{Jizhong Liang}, {and}
  \bibinfo{person}{Yanpeng Li}.} \bibinfo{year}{2015}\natexlab{}.
\newblock \showarticletitle{Improving deep neural networks using softplus
  units}. In \bibinfo{booktitle}{\emph{2015 International Joint Conference on
  Neural Networks (IJCNN)}}. IEEE, \bibinfo{pages}{1--4}.
\newblock


\bibitem[\protect\citeauthoryear{Zhuang, Zhang, Song, and Hooker}{Zhuang
  et~al\mbox{.}}{2021}]%
        {zhuang2021randomness}
\bibfield{author}{\bibinfo{person}{Donglin Zhuang}, \bibinfo{person}{Xingyao
  Zhang}, \bibinfo{person}{Shuaiwen~Leon Song}, {and} \bibinfo{person}{Sara
  Hooker}.} \bibinfo{year}{2021}\natexlab{}.
\newblock \showarticletitle{Randomness in neural network training:
  Characterizing the impact of tooling}.
\newblock \bibinfo{journal}{\emph{arXiv preprint arXiv:2106.11872}}
  (\bibinfo{year}{2021}).
\newblock


\end{thebibliography}
\bibliographystyle{ACM-Reference-Format}

\end{document}